\pdfoutput=1
\documentclass[11pt]{article}

\usepackage[preprint]{acl}

\usepackage{times}
\usepackage{latexsym}

\usepackage[T1]{fontenc}
\usepackage[utf8]{inputenc}
\usepackage{microtype}
\usepackage{inconsolata}

\usepackage{amsmath}
\usepackage{amssymb}
\usepackage{mathtools}
\usepackage{amsthm}
\usepackage{mathrsfs}
\usepackage{nicefrac}
\usepackage{dsfont}
\usepackage{bbm}

\usepackage{graphicx}
\usepackage{booktabs}
\usepackage{multirow, makecell, caption}
\usepackage{subcaption}
\usepackage{tabularx}
\usepackage{colortbl}
\newcolumntype{C}{>{\centering\arraybackslash}X}

\usepackage{enumitem}

\usepackage{hyperref}

\usepackage[capitalize,noabbrev]{cleveref}
\Crefformat{equation}{#2Eq.\;(#1)#3}
\Crefformat{figure}{#2Figure #1#3}

\usepackage{algorithm}
\usepackage{algpseudocode}

\usepackage{listings}
\lstdefinelanguage{json}{
    basicstyle=\ttfamily\small,
    stringstyle=\color{blue},
    morestring=[b]",
    literate=
     *{:}{{{\color{black}:}}}{1}
      {,}{{{\color{black},}}}{1}
}

\usepackage{xspace}
\usepackage{arydshln}
\usepackage{ulem}
\usepackage{wrapfig}

\newcommand{\ie}{{i.e.}} 

\newcommand{\method}{\textsc{HER}\xspace}

\usepackage{color}
\definecolor{mygray}{RGB}{226, 226, 226}
\definecolor{myred}{RGB}{252, 142, 142}
\definecolor{mygreen}{RGB}{147, 255, 143}
\definecolor{myblue}{RGB}{144, 155, 255}
\definecolor{myyellow}{RGB}{253, 253, 143}
\definecolor{mypurple}{RGB}{255, 142, 250}

\ifx\assumption\undefined
\newtheorem{assumption}{Assumption}
\fi

\title{\method: Human-like Reasoning and Reinforcement Learning \\
for LLM Role-playing}

\author{
\small
Chengyu Du$^{\dagger1,2}$  \quad Xintao Wang$^{\dagger1}$ \quad Aili Chen$^{1,2}$ \quad Weiyuan Li$^{1}$ \quad
\textbf{Rui Xu$^{1}$} \quad \textbf{Junteng Liu$^{2}$} \quad \textbf{Zishan Huang$^{2}$} \\
\small
\quad \textbf{Rong Tian$^{2}$} \quad \textbf{Zijun Sun$^{2}$} \quad  \textbf{Yuhao Li$^{2}$} \quad 
\textbf{Liheng Feng$^{2}$} \quad  \textbf{Deming Ding$^{2}$} \quad  \textbf{Pengyu Zhao$^{*2}$} \quad 
\textbf{Yanghua Xiao$^{*1}$} \\[0.5em]
\small
$^{1}$Fudan University \quad $^{2}$MiniMax
}

\begin{document}
\maketitle
\def\thefootnote{$\dagger$}\footnotetext{Project Leader.}\def\thefootnote{\arabic{footnote}}
\begin{abstract}

LLM role-playing, i.e., using LLMs to simulate specific personas, has emerged as a key capability in various applications, such as companionship, content creation and digital games. 
While current models effectively capture character tones and knowledge, simulating the inner thoughts behind their behaviors remains a challenge. 
Towards cognitive simulation in LLM role-play, previous efforts mainly suffer from two deficiencies: 
lacking data with high-quality reasoning traces, 
and lacking reliable reward signals aligned with human preferences. 
In this paper, we propose HER, a unified framework for cognitive-level persona simulation. 
\method introduces dual-layer thinking, which distinguishes characters' first-person thinking from LLMs' third-person thinking. 
To bridge these gaps, we curate reasoning-augmented role-playing data via reverse engineering, and construct human-aligned principles and reward models.
Leveraging these resources, we train \method models based on Qwen3-32B via supervised and reinforcement learning. 
Extensive experiments validate the effectiveness of our approach. 
Notably, our models significantly outperform the Qwen3-32B baseline, achieving a 30.26\% improvement on the CoSER benchmark and a 14.97\% gain on the Minimax Role-Play Bench.
Our datasets, principles, and models are released to facilitate future research~\footnote{https://github.com/cydu24/HER}~\footnote{https://huggingface.co/ChengyuDu0123/HER-32B}~\footnote{https://huggingface.co/ChengyuDu0123/HER-RM-32B}~\footnote{https://huggingface.co/datasets/ChengyuDu0123/HER-Dataset}.

\end{abstract}

\section{Introduction}

\begin{figure}[t]
    \centering
    \includegraphics[width=1.0\linewidth]{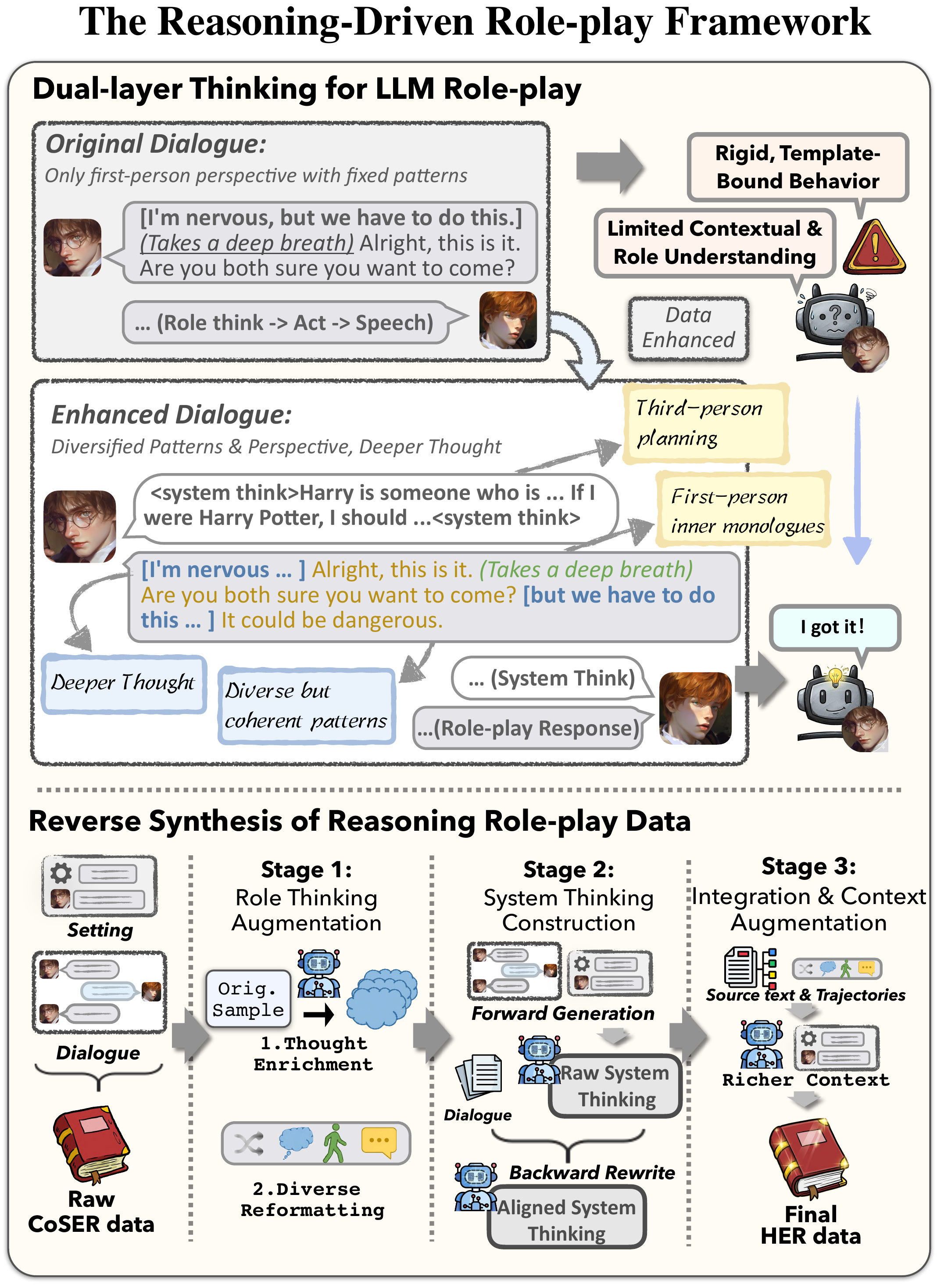}
    \caption{\textbf{The reasoning-driven LLM role-play framework of \textsc{HER}.}
    \textsc{HER} introduces Dual-layer Thinking and a three-stage reverse synthesis pipeline to construct reasoning-augmented LLM role-play trajectories.}
  
    \label{fig:fig1}
\end{figure}

\begin{table*}[t]
\centering
\footnotesize
\setlength{\tabcolsep}{3pt}
\renewcommand{\arraystretch}{0.92}
\setlength{\arrayrulewidth}{0.35pt}
\begin{tabular}{p{0.22\textwidth}|p{0.74\textwidth}}
\hline
\multicolumn{2}{c}{\textbf{Training Sample Format}} \\
\hline
\small\textbf{Input Context} $P, x_{\le t}$ & \textbf{Profile}: \textit{Elizabeth Bennet}---quick-witted, independent, despises false pride. \quad \textbf{Scenario}: Encounters Mr.\ Darcy at Pemberley; previous conflicts unresolved. \\
\hline
\rowcolor{gray!12}
\small\textbf{System Thinking} & \textless system\_thinking\textgreater\ I need to play as ...,My Persona is ...;  Now the Scene is tense reunion... Plan: stay polite, deflect with irony...\ \textless /system\_thinking\textgreater \\
\hline
\small\textbf{Role-play Answer} & \textless role\_thinking\textgreater Why does he look at me so?\textless /role\_thinking\textgreater\ \textless role\_action\textgreater raises eyebrow\textless /role\_action\textgreater\ ``The grounds are unexpectedly pleasant, Mr.\ Darcy.'' \\
\hline
\end{tabular}
\caption{Training sample format. \textbf{Input}: profile + scenario + history. \textbf{Output}: system thinking (3rd-person, hidden) + role-level content (1st-person, visible) evaluated by the GRM. Details in Table~\ref{tab:prompts_agent_2}.}
\label{tab:data_format}
\end{table*}

LLM role-playing, broadly viewed as persona simulation, aims to generate in-character decisions and narratives conditioned on a persona and an evolving scene~\citep{chen2024persona}.  
In this work, we focus on \emph{text-based, multi-turn dialogue} role-playing, where an agent must remain in character throughout an interactive conversation.  
Large language models (LLMs) have demonstrated strong general-purpose language capabilities, largely attributed to large-scale pretraining on natural language corpora, as evidenced by modern frontier LLMs~\citep{deepseekai2025,yang2025qwen3technicalreport}.  
Recently, post-training that emphasizes reasoning and reinforcement learning (RL) has become a central route to further improve model capabilities beyond imitation~\citep{openai2024openaio1card,minimax2025m1}.  
These advances have accelerated role-play applications in companionship, interactive storytelling, and game-like settings, where users expect persistent personas and coherent long-form interactions.  

However, while current LLMs often mimic surface attributes such as speaking style or factual knowledge, \emph{deeply emulating a character’s inner reasoning}---the motivations and plans that connect persona and scene constraints to the next-turn decision---remains challenging.  
The role-playing performance of current reasoning-capable LLMs is still not fully satisfactory and remains to be improved~\citep{liu2025cogdualenhancingdualcognition,ye2025cpoaddressingrewardambiguity}.
Meanwhile, datasets have started to include character inner thoughts, such as CoSER~\citep{coser2024}, but these thoughts are often short and shallow, providing limited supervision for \emph{deep} persona-grounded reasoning.  

Enabling LLMs to deeply simulate inner reasoning in multi-turn dialogue role-play is difficult for two reasons.  
First, although previous efforts have curated role-play dialogues, their underlying planning and inner thoughts are usually implicit, making scalable supervision of reasoning traces costly and inconsistent~\citep{wang2024rolellmbenchmarkingelicitingenhancing,tu2024characterevalchinesebenchmarkroleplaying}.  
For example, the same dialogue turn can be justified by multiple plausible inner motivations, making reasoning annotations inherently subjective and hard to standardize at scale.
Second, role-play outputs are inherently open-ended and non-verifiable, making reward modeling and preference optimization prone to bias and shortcut learning, especially when rewards can be exploited by superficial cues~\citep{liao2025moa}.  
Moreover, rewards can be easily gamed by superficial cues (e.g., verbosity or sentiment words), leading models to optimize for style rather than genuine persona-grounded decisions.

Therefore, effective optimization for reasoning-driven role-play calls for (i) scalable construction of persona-grounded reasoning supervision and (ii) context-dependent reward signals that better align with human preferences in subjective interactions.  

To address these limitations, we propose \textbf{\textsc{HER}} (Human Emulation Reasoning), a unified framework that equips LLMs with structured thinking and trains them with preference-aligned reinforcement learning for role-play.  
\textsc{HER} introduces human-like \textbf{Dual-layer Thinking} (Figure~\ref{fig:fig1}), separating hidden third-person system thinking from supervised first-person inner role thinking.  
We reverse-synthesize reasoning-augmented training data from role-play dialogues in CoSER~\citep{coser2024}, converting each conversation into the Dual-layer Thinking format (Table~\ref{tab:data_format}).

To optimize LLM role-play beyond supervised fine-tuning, we build a self-principled, pair-wise generative reward model, \textbf{HER GenRM}, to provide context-dependent preference signals for role-play optimization.  
We distill a compact set of role-play principles via an expert-alignment workflow and use them to construct preference-style data for training HER GenRM.  
Using HER GenRM as the preference judge, we further train the role-play generator with RL to improve in-character reasoning and decision-making, and validate \textsc{HER} on CoSER Test~\citep{coser2024} and Minimax Role-Play Bench~\citep{MiniMaxAI_roleplay_bench_2026}.

We make three contributions.
(1) We introduce Dual-layer Thinking and a unified training framework for reasoning-driven role-play.
(2) We build \textsc{HER Datasets} with reasoning-augmented trajectories, an expert-distilled principle set, and trained models. We will release these resources for future studies.
(3) We conduct controlled analyses showing distinct gains from system thinking, by-case reward modeling, and balanced anti-shortcut training.

\section{Related Work}

\textbf{LLM role-play and persona simulation.}
Early work explores role-play agents for fictional characters~\citep{shao2023character} and multi-agent simulations of human society~\citep{Park2023GenerativeAgents}, while also analyzing role-play mechanisms~\citep{shanahan2023role} and safety risks~\citep{liu-etal-2024-evaluating-large,deshpande-etal-2023-toxicity}.
Recent surveys~\citep{chen2024from} provide a comprehensive overview of these role-play mechanisms and associated challenges.

\textbf{Role-play datasets.}
Persona datasets, ranging from dialogues~\citep{wang2023rolellm} to multimodal records~\citep{yuan2024evaluating,li2023chatharuhi,dai2024mmrolecomprehensiveframeworkdeveloping}, struggle to balance \emph{authenticity}, \emph{scalability}, and \emph{interaction richness}. 
Synthesized datasets~\citep{wang2023rolellm,chan2024personahub,shao2023character} offer scalability but often lack faithfulness to source materials.
Datasets annotated by humans~\citep{tu2024charactereval,zhou2023characterglm} or extracted from literature~\citep{li2023chatharuhi,chen2023large} ensure authenticity yet are labor-intensive and constrained to simple interaction forms (e.g., QA).
Crucially, most datasets lack explicit \emph{reasoning traces} for character motivations.

\textbf{Role-play evaluation.}
Evaluation methods for RPLAs primarily rely on multiple-choice benchmarks to assess isolated facets~\citep{shen2023roleeval,xu2024character,yuan2024evaluating,wang2024incharacter} or LLM judges for open-ended dimensions~\citep{tu2024charactereval,zhou2023characterglm,wang2023rolellm,shao2023character}.
However, these benchmarks lack interaction dynamics, and LLM judges suffer from biases~\citep{li2024judgesurvey}. Consequently, current methods fail to balance subjective nuances with factual consistency, directly impeding Reward Modeling (RM), which requires robust signals for optimization.

\textbf{Reasoning in role-play alignment.}
Early alignment approaches primarily relied on Supervised Fine-Tuning (SFT)~\citep{coser2024,wang2025opencharacter} for stylistic imitation.
Following reasoning models like OpenAI-o1~\citep{openai2024openaio1card} and DeepSeek-R1~\citep{deepseekr1}, the paradigm shifted toward RL optimization via GRPO~\citep{deepseekgrpo} and DAPO~\citep{yu2025dapo}.
In the role-play domain, techniques like MOA~\citep{liao2025moa}, RAIDEN-R1~\citep{wang2025raidenr1}, and CogDual~\citep{liu2025cogdual} leverage reasoning to improve role-play consistency.
However, they rely on verifiable keywords or static principles, which fail to offer context-dependent adaptability and remain vulnerable to exploitable shortcuts.
Alignment frameworks~\citep{ye2025cpo} or affective alignment~\citep{zhang2025echon1} focus on surface-level output styles, neglecting the alignment of internal reasoning with a character's unique logic.


\section{Method}
\label{sec:method}
\begin{figure*}[t]
    \centering
    \includegraphics[width=1.0\linewidth]{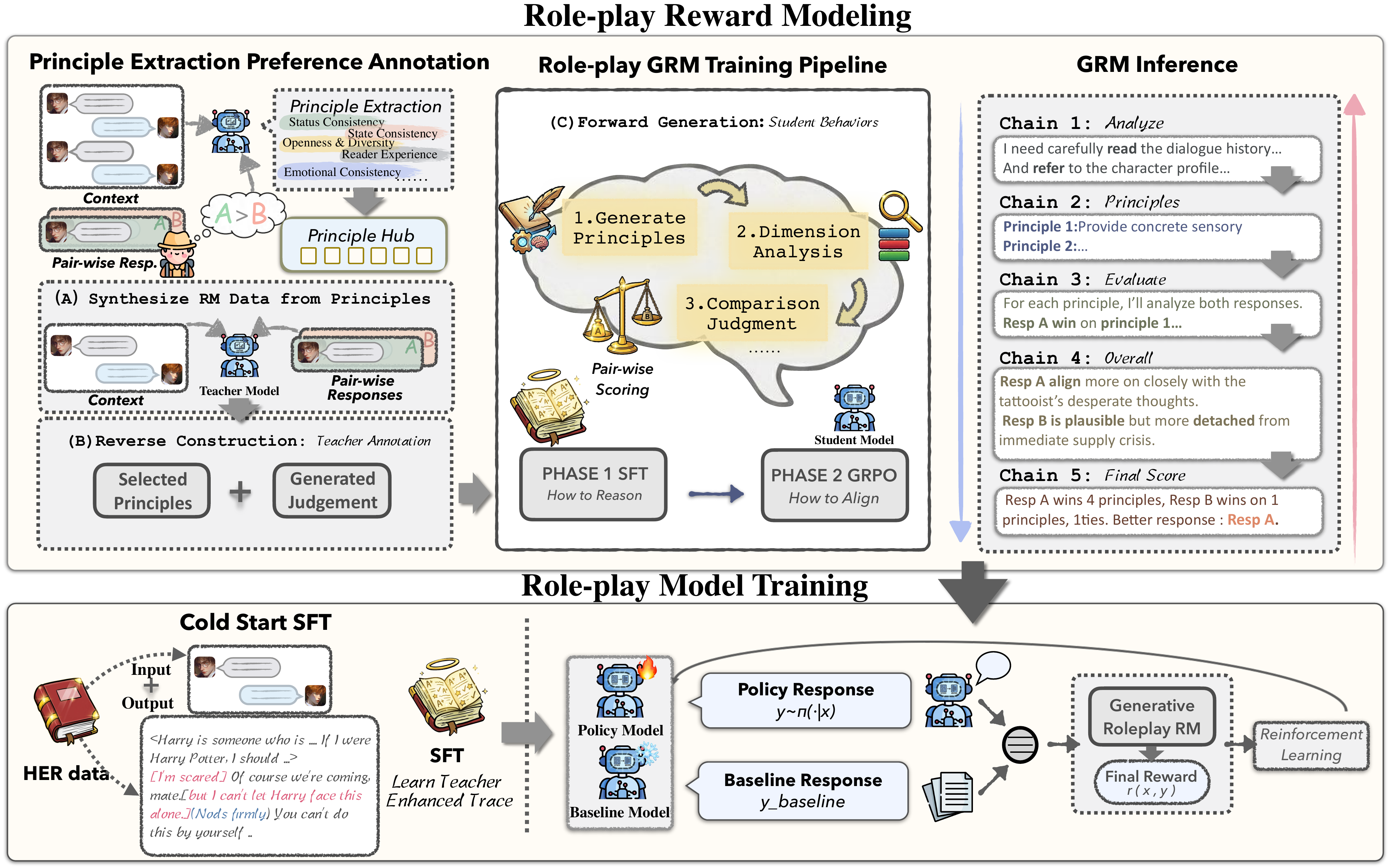}
    \caption{\textbf{Overview of \textsc{HER} training.}
\textbf{Top:} we train a Role-play GRM by distilling reusable principles from real conversational preference data, and teaching the model to do pairwise judging with \emph{by-case principles} $\rightarrow$ analysis $\rightarrow$ final decision. \textbf{Bottom:} we first cold-start the LLM role-play model with SFT on \textsc{HER} data, and then apply RL where the GRM compares the policy response with a baseline response to produce the reward.
}
    \label{fig:fig2}
\end{figure*}

We aim to improve LLM role-play by making the model \emph{think before it speaks}, and then optimize this behavior using reinforcement learning.
Our method has four parts: Dual-layer Thinking~\S\ref{sec:thinking}, reasoning data synthesis~\S\ref{sec:data}, a principle-aligned Role-play GRM (Generative Reward Model)~\S\ref{sec:rm}, and RL for the LLM role-play generator~\S\ref{sec:rl}.

\subsection{Dual-layer Thinking}
\label{sec:thinking}

\textbf{Why Dual-layer Thinking is necessary.}
\method{} introduces Dual-layer Thinking to define an output format for reasoning-capable LLM role-play models.  

In many tasks, \emph{thinking} is a hidden reasoning process, while the \emph{answer} is the user-facing output.  
Role-play needs both: a hidden planner to track constraints, and visible inner thoughts that make the character believable.  
We call the third-person planning process \textbf{system thinking}, and the character’s first-person inner thoughts \textbf{role thinking}.  

System thinking happens before the response, is hidden from users, and is never exposed to the GRM or the evaluator.
Its role is to spend more computation on understanding the persona and scene constraints, and planning the next-turn trajectory.  
Role thinking is part of the LLM role-play transcript and is supervised by the reward model as content.  
Role thinking models the character’s internal state---including emotions, intentions, and decisions---right before generating visible speech or actions.  
In LLM role-play, these first-person thoughts are exactly what users care about, so hiding them inside system thinking makes the training target incomplete.  

Existing methods often fail to distinguish system reasoning from role thinking~\citep{tang2025thinkingcharacteradvancingroleplaying}. This conflation causes two issues: (1) the lack of a dedicated planner for following role constraints, and (2) the inability of reward models to supervise role thinking when mixed with system thinking specifically. Our Dual-layer Thinking in Figure~\ref{fig:fig2} decouples these processes by generating system thoughts first, allowing subsequent role outputs to interleave thoughts with actions and speech.

\textbf{Formal definition}
We model each dialogue turn $t$ as a two-stage generation process.  

Given the dialogue history $x_{\leq t}$ and the global conversation setting $S$, the model first produces a third-person system thinking $s_t$:  
\begin{equation}
\small
s_t \sim \pi_\theta^{\text{sys}}(\cdot \mid S, x_{\leq t}).
\end{equation}
Conditioned on $s_t$, the model then generates role-level outputs:  
\begin{equation}
\small
y_t = (e_1, e_2, \ldots, e_{K_t}) \sim \pi_\theta^{\text{role}}(\cdot \mid S, x_{\leq t}, s_t),
\end{equation}
where each element $e_k \in  \mathcal{R} \cup \mathcal{A} \cup\mathcal{U}$ is role thinking $r$, action $a$, or speech $u$.  

The ordering and composition of $\{e_k\}$ are decided by the model based on context rather than a fixed template.  
In the final transcript, role-level elements are visible, while system thinking is hidden; however, role thinking is evaluated as part of the answer space.  
System thinking and role thinking can be displayed or collapsed based on application designs. Following~\citep{deepseekr1}, we only discard previous system thinking for multi-turn conversations.

\subsection{Reasoning Data Synthesis for LLM Role-play}
\label{sec:data}

High-quality, human-written LLM role-play dialogues are widely available in novels and online communities, but their underlying reasoning is usually implicit.
While human readers can often infer the character's hidden thoughts and motivations from the dialogue---a process that inspires our reverse synthesis---manual annotation is expensive and hard to scale.
We therefore propose an automated, LLM-driven reverse-engineering pipeline to reconstruct both system thinking and role thinking from surface dialogues.
This pipeline converts existing LLM role-play dialogues into large-scale, reasoning-augmented trajectories without manual effort.
We leverage the commercial model as a teacher model to collect high-quality datasets.
Dual-layer Thinking requires training data that contains both system thinking and role thinking, so synthesis is necessary.
We use mutually disjoint splits for LLM role-play SFT, GRM training, policy RL, and evaluation to prevent data leakage; the construction protocol and statistics are reported in Appendix~\ref{app:data_splits}.

\textbf{Input and output}
Each raw sample provides an LLM role-play prompt $P$ (persona card + scenario) and a multi-turn dialogue $x_{1:T}$.  

Our goal is to output a trajectory where each turn has one hidden system thinking and a role-level sequence that interleaves role thinking, role action, and speech.  
We build this trajectory with a three-stage synthesis pipeline.  

\textbf{Stage 1: Role thinking augmentation}
Stage 1 synthesizes first-person role thinking to explain the character's next action or utterance.

\textbf{(1) CoT synthesis}
A strong teacher model generates role thinking to state emotions and intentions. In the same pass, we revise the paired role action and speech to enforce within-turn consistency.
\textbf{(2) Diversity reformatting}
We rewrite each turn into multiple layouts by varying the interleaving of thoughts, actions, and speech.
This balances common structures (e.g., think$\rightarrow$speak vs.\ think$\rightarrow$act$\rightarrow$speak), increasing unique patterns (661$\to$939) and preventing template collapse (Appendix~\ref{app:diversity_pattern}).

\textbf{Stage 2: System thinking construction}
Stage 2 constructs third-person system thinking that plans the next-turn trajectory.
(1) \textbf{Forward generation}
We first generate a draft plan from the current prompt $P$ and history $x_{\le t}$.
(2) \textbf{Backward rewrite}
The forward draft can mix viewpoints or miss what the character actually does next, because it has not seen the true continuation.  
We therefore rewrite it using the ground-truth continuation.
This rewrite removes first-person inner thoughts from system thinking to avoid mixing it with role thinking.  
We also test system thinking effect with a direct ablation in Section~\ref{sec:system_thinking_ablation}.  

\textbf{Stage 3: Integration \& Context augmentation}
Stage 3 repairs the LLM role-play system prompt $P$ so that it faithfully supports the synthesized reasoning and reduces hallucinations in later turns.  
Since the original prompt may lack constraints for the richer synthesized content, we cross-check it against the source novel and current dialogue.
We add missing facts and remove unsupported details while keeping the original meaning, which provides explicit constraints for the GRM to learn valid by-case principles.
Based on this pipeline, we construct the \textsc{HER} dataset (refer to Appendix~\ref{app:data_synthesis}).

\subsection{Role-play GRM}
\label{sec:rm}

To improve an LLM role-play generator with RL, we need a reward model that can tell which response is better.  
This is hard for LLM role-play because responses are open-ended and there is no single verifiable answer.  

We learn the reward signal from \textbf{real preferences}, so that the reward model can mimic how real humans judge and rank LLM role-play responses.  
Specifically, we train a \textbf{Role-play GRM}, a generative reward model that produces an evaluation trace and a final preference for a response pair.  

The GRM performs \textbf{pairwise} comparison and outputs $y \in \{\texttt{cand\_1}, \texttt{cand\_2}, \texttt{tie}\}$.  
In this setting, RL is only as good as the reward model that provides its training signal. Details in Table~\ref{tab:prompts_genrm}.

\textbf{Design}
Instead of scoring with a single number, our GRM follows a process: (1) generating \textbf{by-case principles} to capture scene-specific implicit preference constraints according to the dialogue; (2) analyzing candidates against these principles with concrete pros and cons; and (3) outputting a \textbf{binary preference} (or tie).
This design makes the reward signal both context-sensitive and checkable.

\textbf{Principles distillation}
We distill a compact set of principles from high-quality LLM role-play interaction patterns.
Concretely, a teacher LLM analyzes 300k simulated preference pairs and generates 3--5 principles per pair, producing 36,373 unique raw principles.
We cluster them into 15 semantic categories and select high-frequency representatives, resulting in 107 candidate principles.
Domain experts then merge redundancies, clarify wording, and fill missing criteria, yielding 51 finalized principles across 12 dimensions.
The resulting set reflects interaction-driven criteria rather than benchmark-specific principles (Appendix~\ref{app:principle_distill}).

\textbf{Preference Data Synthesis}
To build GRM training data, we sample a dialogue context $x$ and generate two candidates $A$ and $B$ from a base LLM role-play model.
A strong teacher judge then uses the principle set as a reference library to produce (i) selected principles, (ii) structured analysis, and (iii) the final label $y^*$.
We audit the teacher-labeled preferences on a held-out expert-annotated set disjoint from both training and evaluation data, obtaining \textbf{80.5\%} agreement with expert consensus (Appendix~\ref{app:human_alignment}).

\textbf{Training: SFT then RL}
We train the GenRM in two stages using the synthesized preference data: SFT to learn the full judging trajectory (principles, analysis, and verdict), and then RL to improve verdict correctness. The reward for the RL stage is defined as $R(\hat{y}, y^*) = 1$ if $\hat{y}=y^*$ and $-1$ otherwise. To prevent shortcut learning, we balance candidate order and mix judging formats (Appendix~\ref{app:pattern_parsing}).

\textbf{Balanced construction to reduce shortcuts}
During GenRM training, unbalanced data can induce shortcut behaviors such as position bias, length bias, or collapsing into one fixed judging template; we therefore balance candidate order, include length-contrastive pairs, and mix multiple judging formats. 
We also explain why we use pairwise judgment for GRM rather than point-wise (details in Section~\ref{sec:rm_analysis}).

\subsection{Reinforcement Learning for Role-play Generation}
\label{sec:rl}

With the trained GenRM as a judge, we further improve the LLM role-play generator beyond SFT using outcome-based RL with a clipped policy objective.
We initialize the policy model $\pi_\theta$ from the SFT checkpoint and keep the SFT model frozen to produce a baseline response for comparison.
The frozen SFT response serves as a stable baseline.

\textbf{Reward from pairwise comparison}
For each context $x$, we sample a response $y \sim \pi_\theta(\cdot \mid x)$ and pair it with a baseline response $y_{\text{sft}}$ generated by the frozen SFT model.
The GenRM judges $(x, y, y_{\text{sft}})$ and we parse its final verdict (\texttt{win/lose/tie}) using rules, which is mapped to a scalar reward: $r(x,y)=1$ if $y \succ y_{\text{sft}}$; $-1$ if $y \prec y_{\text{sft}}$; and $0$ if $y \approx y_{\text{sft}}$.
This completes a closed loop: Dual-layer Thinking defines what to model, the GRM defines what to reward, and RL pushes the generator toward stable, in-character LLM role-play. Optimization details in Appendix~\ref{app:rl_details}.


\section{Experiments}
\label{sec:experiments}

\subsection{Experimental Setup}
\label{sec:exp_setup}

\textbf{Benchmarks}
We use the CoSER benchmark~\citep{coser2024} as the main benchmark for multi-turn LLM role-play quality. We use the official CoSER prompts and scoring principle, and format the model outputs into a unified tag-based transcript for evaluation (Appendix~\ref{app:eval_prompts}). 
CoSER reports four scores: Story Consistency (SC), Anthropomorphism (AN), Character Fidelity (CF), and Storyline Quality (SQ).  
We include Minimax Role-Play Bench~\citep{MiniMaxAI_roleplay_bench_2026} as a 100-turn self-chat and follow its official protocol. Full settings and scoring details are in Appendix~\ref{app:long_dialog}.

\textbf{Metrics}
For CoSER Test, we report the average score and four dimension scores (SC/AN/CF/SQ). 
For Minimax Role-Play Bench, we report Worlds, Stories, Preferences, three dimensions and sub-dimensions.

\textbf{Models}
We compare \textsc{HER} with strong commercial LLMs and open-source baselines.  
\textbf{HER} is trained on Qwen3-32B-Base~\citep{qwen3technicalreport}.

\textbf{Evaluation protocol}
We follow CoSER Test~\citep{coser2024} and evaluate 200 held-out conversations, each containing 20 rounds.  
For each conversation, we use an LLM judge to score four dimensions (SC/AN/CF/SQ) \emph{point-wise} on a 100-point deduction-based principle.  
We use \texttt{Qwen3-235B-A22B}~\citep{qwen3technicalreport} as the judge for CoSER evaluation.  
All compared systems are prompted to produce the same LLM role-play transcript format (role thinking/action/speech tags); the exact prompts and principles are provided in Appendix~\ref{app:eval_prompts}.  
For models that generate, we remove the system thinking in previous turns. 
All open source models are decoded with temperature 0.7 and max tokens 4096.

\textbf{Human--LLM alignment (training labels)}
On the held-out 50 cases, the agreement between expert consensus and teacher preference labels reaches \textbf{80.5\%}; disagreements mainly involve subtle emotional expression and subjective style preferences (Appendix~\ref{app:human_alignment}).
For benchmark evaluation, we validate the CoSER judge via expert calibration and further confirm inter-judge consistency on induced pairwise preferences (Appendix~\ref{app:judge_robustness}).

\subsection{Main Results}
\label{sec:main_results}

Table~\ref{tab:main_leaderboard} reports results on CoSER (main), and Minimax Role-Play Bench.  
On CoSER, \textbf{HER-RL} achieves 53.1 average score, outperforming CoSER-70B (35.8) by +17.3 points and \emph{narrowing the gap} to commercial models (Table~\ref{tab:main_leaderboard}).  
HER-RL improves over HER-SFT (53.1 vs.\ 50.9), showing that RL brings gains beyond SFT.  
HER-RL achieves a large gain in Storyline Quality (SQ: 58.1), matching our goal of improving long-range plot progression.  
On Minimax Role-Play Bench, HER-RL scores 65.7, significantly outperforming the SFT baseline (58.4) and CoSER-70B (45.4), particularly in user interaction preference (86.9 vs.\ 82.6) as shown in Table~\ref{tab:main_leaderboard}.

\begin{figure}[h]
    \centering
    \includegraphics[width=1\linewidth]{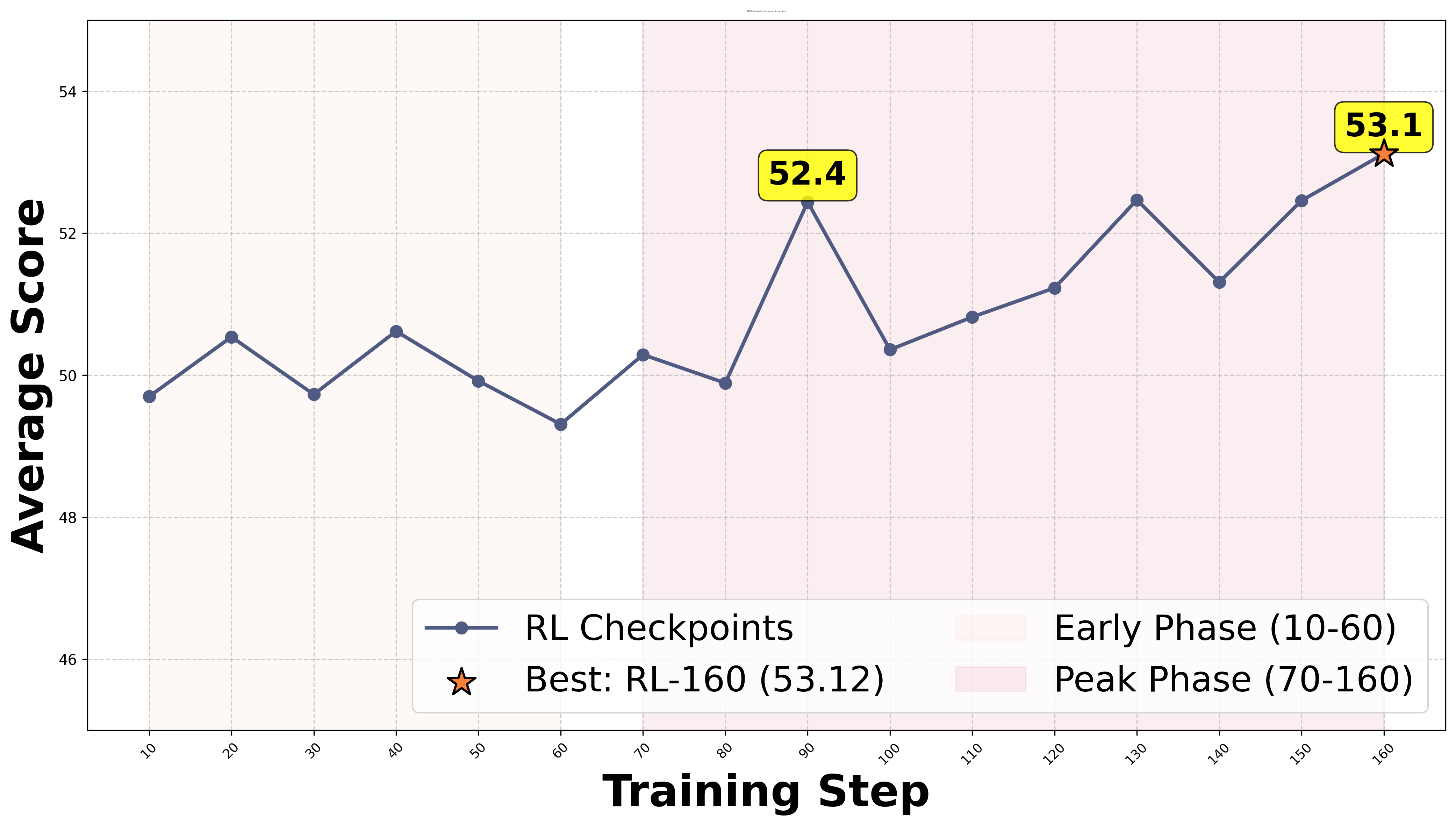}
    \caption{\textbf{Performance of \textsc{HER} Role-play RL training on CoSER Benchmark.}}
    \label{fig:rl_ablation_avg_score}
\end{figure}
\definecolor{coser-blue}{rgb}{0.92, 0.96, 1.0}        
\definecolor{basic-yellow}{rgb}{0.98, 0.96, 0.90}     
\definecolor{adv-green}{rgb}{0.92, 1.0, 0.92}         
\definecolor{know-purple}{rgb}{0.95, 0.90, 0.95}      

\definecolor{row-ours}{rgb}{0.95, 1.0, 0.95}          
\definecolor{row-her}{rgb}{0.95, 0.98, 1.0}           
\definecolor{row-tier1}{rgb}{0.98, 0.98, 0.98}        
\definecolor{row-ext}{rgb}{0.95, 0.95, 0.95}          

\definecolor{sage-green}{rgb}{0.447, 0.561, 0.431}    
\definecolor{soft-coral}{rgb}{0.976, 0.835, 0.804}    
\definecolor{sky-blue}{rgb}{0.804, 0.878, 0.969}      

\begin{table*}[h]
    \centering
    \small
  
    \resizebox{\textwidth}{!}{
    \begin{tabular}{clccccc|ccccc}
    \toprule

    \multirow{3}{*}{\textbf{Rank}} &
    \multirow{3}{*}{\textbf{Model}} &
    \multicolumn{5}{c|}{\cellcolor[rgb]{0.92, 0.96, 1.0}\textbf{CoSER Benchmark}} &
    \multicolumn{5}{c}{\cellcolor[rgb]{0.98, 0.94, 0.90}\textbf{Minimax Role-Play Bench}} \\

    \cmidrule(lr){3-7} \cmidrule(lr){8-12}

    & &
    \multirow{2}{*}{\textbf{Avg}} & \multirow{2}{*}{\textbf{SC}} & \multirow{2}{*}{\textbf{AN}} & \multirow{2}{*}{\textbf{CF}} & \multirow{2}{*}{\textbf{SQ}} &
    \multirow{2}{*}{\textbf{Avg}} & \textbf{Worlds} & \textbf{Stories} & \textbf{Pref} & \multirow{2}{*}{\textbf{95\% CI}} \\

    & & & & & & &
    & \textbf{(50\%)} & \textbf{(25\%)} & \textbf{(25\%)} & \\

    \midrule

    \rowcolor[rgb]{0.98, 0.98, 0.98}
    1 & Claude-4.5-Opus & \textbf{62.43} & 63.74 & \textbf{64.28} & 58.45 & 63.24 & 76.62 & 67.23 & 82.10 & 89.90 & [75.5, 77.7] \\

    \rowcolor[rgb]{0.98, 0.98, 0.98}
    2 & Gemini-3-Pro & 61.80 & \textbf{65.95} & 60.42 & \textbf{58.34} & 62.49 & 75.60 & 62.72 & 83.87 & 93.08 & [74.5, 76.7] \\

    \rowcolor[rgb]{0.98, 0.98, 0.98}
    3 & GPT-5.1 & 61.10 & 64.95 & 53.99 & 60.13 & 65.35 & 80.63 & 76.62 & 72.21 & 97.05 & [79.6, 81.6] \\

    \rowcolor[rgb]{0.98, 0.98, 0.98}
    4 & Gemini-2.5-Pro & 60.68 & 61.05 & 60.80 & 57.48 & 63.40 & 68.23 & 52.36 & 82.11 & 86.08 & [67.1, 69.3] \\

    \midrule

    5 & DeepSeek-v3.2 & 58.68 & 55.85 & 57.07 & 57.44 & 64.35 & 60.27 & 45.81 & 66.64 & 82.83 & [59.2, 61.4] \\

    6 & \textbf{MiniMax-M2-her} & 57.30 & 60.03 & 50.11 & 49.30 & \textbf{69.77} & \textbf{84.65} & \textbf{80.55} & 79.97 & \textbf{97.51} & [83.6, 85.7] \\

    7 & DeepSeek-v3.1 & 53.50 & 50.15 & 53.18 & 53.93 & 56.72 & 64.22 & 51.11 & 66.45 & 88.21 & [62.9, 65.5] \\

    \rowcolor[rgb]{0.95, 0.98, 1.0}
    8 & \textbf{HER-RL} & 53.12 & 54.33 & 47.26 & 52.78 & 58.12 & 65.73 & 59.13 & 57.74 & 86.90 & [63.0, 68.4] \\

    \rowcolor[rgb]{0.95, 0.98, 1.0}
    9 & \textbf{HER-SFT} & 50.92 & 50.52 & 45.99 & 49.78 & 57.37 & 58.44 & 47.29 & 52.78 & 86.40 & [56.5, 60.4] \\

    \midrule

    10 & Grok-4.1-Fast & 47.40 & 49.21 & 47.57 & 42.64 & 50.17 & 48.47 & 29.87 & 47.51 & 86.64 & [47.4, 49.5] \\

    \rowcolor[rgb]{0.98, 0.98, 0.98}
    11 & Claude-4.5-Sonnet & 45.21 & 47.18 & 36.02 & 47.55 & 50.09 & 69.35 & 55.72 & 75.66 & 90.28 & [68.2, 70.5] \\

     \midrule
    12 & Claude-3.7-Think & 39.73 & 44.84 & 31.00 & 42.45 & 40.65 & 61.25 & 50.66 & 59.53 & 84.15 & [58.5, 64.0] \\

    13 & CoSER-70B & 35.95 & 35.05 & 31.16 & 32.28 & 45.33 & 45.38 & 34.32 & 30.32 & 82.58 & [43.5, 47.2] \\

    14 & GPT-5-Mini & 32.97 & 38.10 & 24.60 & 27.20 & 42.00 & 57.63 & 43.32 & 50.11 & 93.78 & [55.9, 59.3] \\
 
     \midrule
    15 & GPT-4o-240806 & 27.69 & 34.00 & 14.90 & 22.90 & 38.90 & 66.39 & 64.96 & 46.23 & 89.40 & [64.1, 68.7] \\

    16 & GPT-OSS-120B & 26.12 & 32.80 & 14.80 & 21.50 & 35.40 & 60.72 & 47.27 & 56.65 & 91.71 & [58.0, 63.4] \\

    17 & Qwen3-32B & 22.86 & 30.56 & 19.61 & 15.52 & 30.56 & 50.76 & 40.38 & 32.82 & 89.48 & [48.4, 53.2] \\

    \bottomrule
    \end{tabular}
    }
    \vspace{0.5em}
      \caption{\textbf{Main Leaderboard: CoSER \& Minimax Role-Play Bench.}
    CoSER: 0-100 (higher is better), evaluating story consistency and character fidelity.
    MiniMax: 0-100, evaluating worlds (50\%), stories (25\%), and preferences (25\%).
    Full results in Table~\ref{tab:minimax_new_category}.}
    \label{tab:main_leaderboard}
\end{table*}


\begin{table}[t]
\centering
\footnotesize
\setlength{\tabcolsep}{8pt}
\begin{tabular}{lc}
\toprule
Format (additive ablation) & Agreement \\
\midrule
General principles + point-wise (no CoT)         & 60.0\% \\
By-case principles + point-wise (no CoT)         & 86.0\% \\
By-case principles + pairwise (no CoT)           & 88.0\% \\
By-case principles + pairwise (+CoT)             & \textbf{93.0\%} \\
\bottomrule
\end{tabular}
\caption{\textbf{GRM supervision format on 5k preferences.}
By-case principles are generated from preference context}
\label{tab:rm_format}
\end{table}

\subsection{Reward Model Supervision: General vs.\ By-case Principles}
\label{sec:rm_analysis}

We compare by-case principles with fixed principles on a test set of 4,739 preference pairs annotated by human experts.
All GRM variants in this section are trained from the same SFT checkpoint; only the supervision format differs.
Further details on data construction are in Appendix~\ref{app:pair_data_construction}.

Table~\ref{tab:rm_format} shows a reward model must align with the role-play context; otherwise, fixed expert-written principles can miss what matters to a specific character in a specific scene.

We evaluate with the agreement ratio, i.e., whether the GRM prefers the same response as human experts. As shown in Table~\ref{tab:rm_format}, GRM with fixed principles reaches 60.0\% agreement, while GRM with by-case principles achieves 86.0\%.
Under the same by-case principles, pairwise comparison further improves agreement from 86.0\% to 88.0\%, since independent point-wise scoring is harder to calibrate across candidates.  
Adding CoT in the analysis trace increases agreement from 88.0\% to 93.0\%, so we adopt \textbf{by-case principles + pairwise + CoT} as the final GRM supervision format.

\subsection{Preventing Reward Hacking with Balanced Data}
\label{sec:reward_hacking}

Even strong reward models can be easily reward-hacked under imbalanced preference pairs; beyond position and length biases, we focus on a mode unique to multi-dimensional judging: \emph{pattern bias}.  

Our GRM (trained on Qwen3-32B) evaluates each response pair by first generating multiple \emph{by-case principles} (dimensions), then comparing the two candidates under each principle, and finally producing an overall \texttt{win/lose/tie} decision.  

\begin{figure}[h]
    \centering
    \includegraphics[width=1.0\linewidth]{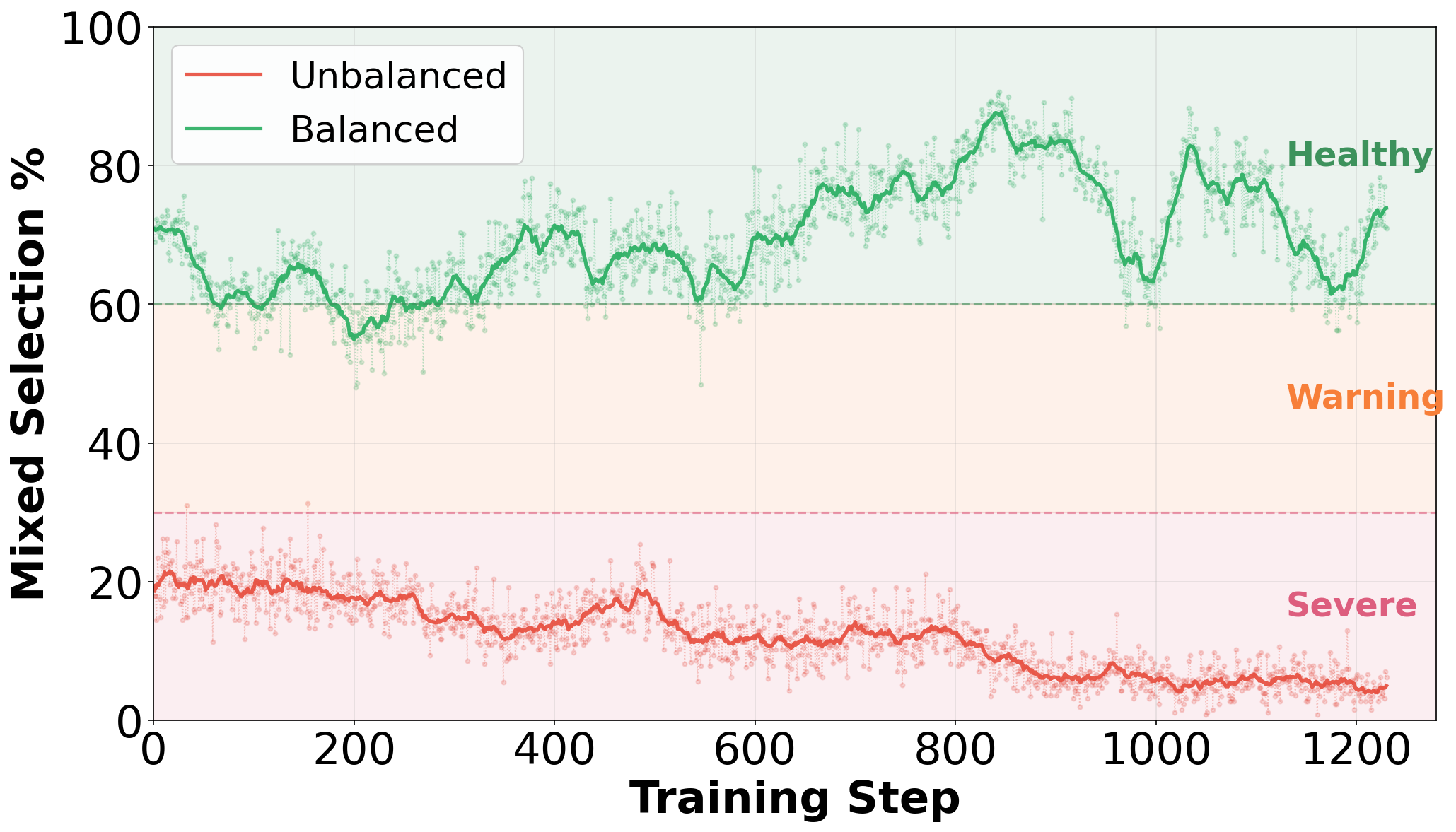}
    \caption{\textbf{Pattern collapse vs.\ stable dimension-wise judgments during GRM RL training.}}
    \label{fig:pattern_bias_trends}
\end{figure}

We call it \emph{pattern bias} when dimension-wise comparisons collapse into ``all-A'' or ``all-B'', \ie, the model claims the same side wins on \emph{every} principle.  
We quantify this shortcut using \emph{Mixed Selection \%}: the dimension-wise winners are \emph{not} uniform (i.e., not all-A and not all-B), meaning the judge considers trade-offs across dimensions rather than assigning every dimension to one side.  
As shown in Figure~\ref{fig:pattern_bias_trends}, under an unbalanced mixture the GRM quickly collapses toward uniform patterns (18.0\%$\rightarrow$6.2\%), while the balanced mixture maintains a stable non-uniform rate (69.0\%$\rightarrow$70.9\%).  
\begin{table}[t]
\centering
\small
\begin{tabular}{lcc}
\toprule
\textbf{Metric} & \textbf{Unbalanced} & \textbf{Balanced} \\
\midrule
\multicolumn{3}{l}{\textit{Pattern Bias (GRM Training Dynamics)}} \\
\quad Mixed Selection (Start) & 18.0\% & 69.0\% \\
\quad Mixed Selection (End) & 6.2\% & 70.9\% \\
\quad Pattern Bias (End) & 93.8\% & 29.1\% \\
\midrule
\multicolumn{3}{l}{\textit{GRM Quality (Test Set, N=800)}} \\
\quad Accuracy & 69.91\% & \textbf{73.99\%} \\
\quad $\Delta$ vs Unbalanced & — & +4.08\% \\
\bottomrule
\end{tabular}
\caption{Comparison of balanced vs.\ unbalanced GRM RL training on Qwen3-32B.
}
\label{tab:pattern_bias_comparison}
\end{table}
We mitigate this shortcut by balancing training construction and mixing different judging patterns with controlled proportions in Appendix~\ref{app:pattern_parsing}.


\subsection{System Thinking Improves Character Fidelity}
\label{sec:system_thinking_ablation}
\begin{figure}[t]
    \centering
        \includegraphics[width=0.7\linewidth]{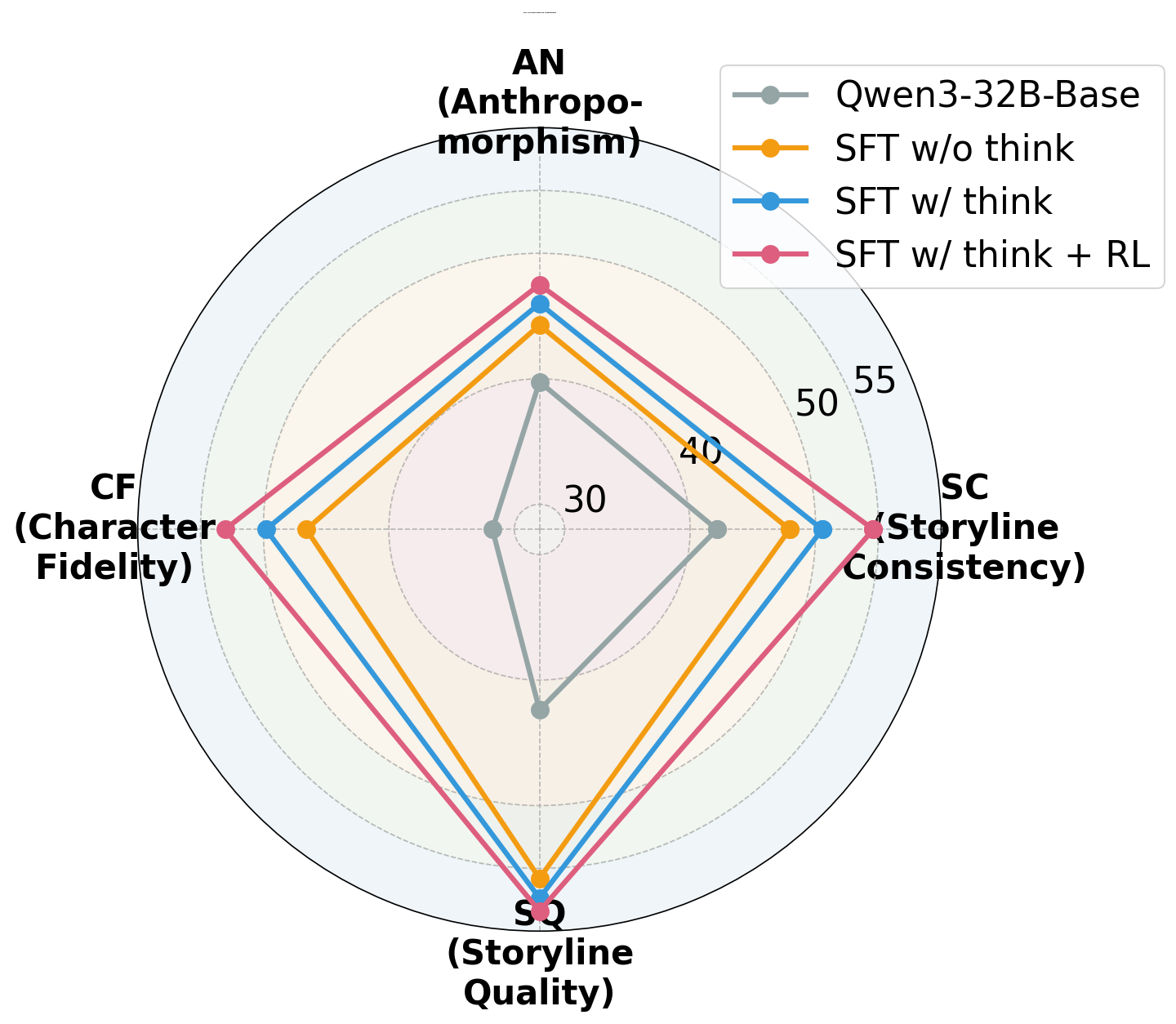}
    \caption{\textbf{Effect of system thinking and RL on CoSER Benchmark.}
We compare a base model, SFT without thinking, SFT with system\_thinking, and RL model.}
    \label{fig:system_thinking_ablation}
\end{figure}

\begin{table}[t]
\centering
\small
\begin{tabular}{lclc}
\toprule
\multicolumn{2}{c}{\textit{Statistics}} & \multicolumn{2}{c}{\textit{Distribution}} \\
\midrule
Mean & 580 tokens & $<$250 tokens & 15.0\% \\
Min & 77 tokens & 250--500 & 42.5\% \\
Max & 1,443 tokens & 500--750 & 30.0\% \\
Range & 18$\times$ & $>$750 tokens & 12.5\% \\
\bottomrule
\end{tabular}
\caption{\textbf{System thinking length statistics.}}
\label{tab:thinking_length}
\end{table}
\begin{table}[t]
\centering
\small
\begin{tabular}{lcc}
\toprule
\textbf{Metric} & \textbf{Collapsed} & \textbf{Diversified} \\
\midrule
\multicolumn{3}{l}{\textit{Structure-level diversity}} \\
Top-1 Pattern (\%) $\downarrow$ & 96.1 & \textbf{49.2} \\
Unique Patterns $\uparrow$ & 4 & \textbf{21} \\
Shannon Entropy $\uparrow$ & 0.28 & \textbf{2.15} \\
\midrule
\multicolumn{3}{l}{\textit{Token-level diversity}} \\
Distinct-2 $\uparrow$ & 0.4329 & 0.4256 \\
Distinct-4 $\uparrow$ & 0.8743 & 0.8677 \\
\midrule
\multicolumn{3}{l}{\textit{Cross-sample similarity (Self-BLEU)}} \\
Self-BLEU (2-gram) $\downarrow$ & 0.0392 & \textbf{0.0237} \\
Self-BLEU (4-gram) $\downarrow$ & 0.0140 & \textbf{0.0013} \\
\bottomrule
\end{tabular}
\caption{Diversity metrics comparison. Self-BLEU measures cross-sample n-gram overlap (lower = more diverse). The gap widens with longer n-grams (11$\times$ at 4-gram), indicating Collapsed outputs share more long repeated phrases.}
\label{tab:diversity_comparison}
\end{table}

We test whether enabling explicit system thinking during training and inference improves in-character ability.
Specifically, the model generates an explicit system thinking block before each response to reason about character traits and response strategy.
This system thinking is generated only for the current turn, is not appended to the dialogue history, and is removed before evaluation.
As shown in Figure~\ref{fig:system_thinking_ablation}, enabling system thinking improves the average score from 48.64 to 50.92, with the largest gains observed on Character Fidelity (+3.21) and Storyline Consistency (+2.60).
Applying RL on top of system thinking further boosts the average score to 53.12, with improvements again concentrated on consistency-related dimensions.
The model also adaptively adjusts thinking length based on the scenario, as shown in Table~\ref{tab:thinking_length}.  


\subsection{Keeping Diversity During Role-play Training}
\label{sec:diversity_ablation}

LLM role-play also benefits from diverse \emph{response structures} (how thoughts, actions, and speech are interleaved); otherwise, training can collapse into a single pattern that produces less expressive interactions.  
We improve diversity by rewriting SFT trajectories to mix multiple valid interleaving patterns of role thinking, actions, and speech.

\begin{figure}[H]
    \centering

    \includegraphics[width=1.0\linewidth]{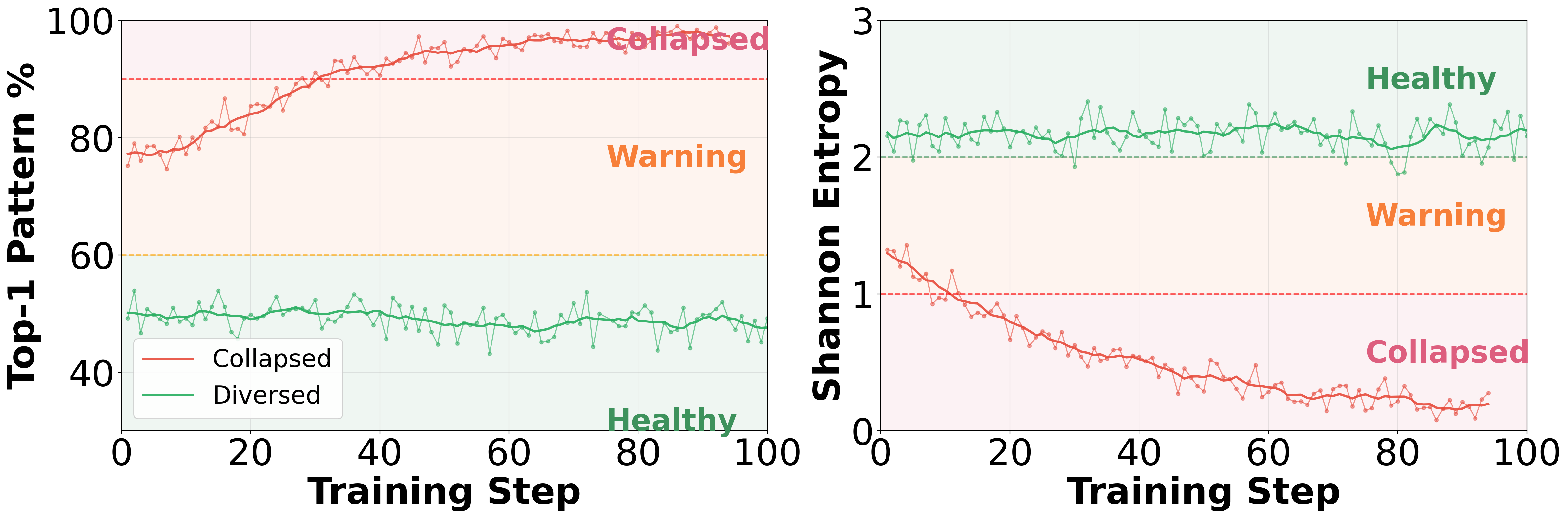}
    \caption{\textbf{Pattern collapse vs.\ stable diversity during RL training.}
The \textbf{Collapsed} run quickly concentrates on a single pattern (Top-1 rises above 90\% and entropy drops), while the \textbf{Normal} run stays below the collapse threshold and maintains substantially higher Shannon entropy.}
    \label{fig:collapse_trend}
\end{figure}

Figure~\ref{fig:collapse_trend} shows the collapse dynamics: in the \textbf{Collapsed} setting, Top-1 pattern concentration crosses the 90\% threshold by step~28 and reaches 96.3\% at step~50 with entropy dropping from 1.32 to 0.29; in contrast, the \textbf{Diversified} setting maintains Top-1 concentration between 43--54\% throughout 100 steps and keeps entropy consistently above 2.0. 
Details in Appendix~\ref{app:diversity_pattern}.

\section{Conclusion}
We study how to train large language models to \emph{think in character} for role-play.
We introduce \textsc{HER}, a unified framework combining Dual-layer Thinking, three-stage reverse synthesis, a principle-aligned Role-play GRM, and RL.
Experiments on CoSER show gains in character fidelity and narrative quality.
We hope \textsc{HER} offers a reproducible path to role-play models that \emph{think} in character.

\newpage
\section*{Limitations}
We discuss several limitations of our work. First, our evaluation is primarily based on the CoSER benchmark, which may not capture all aspects of roleplay quality. Second, our reasoning-aware data construction relies on strong teacher models, which introduces computational costs. Third, while we analyze position and length biases, other forms of reward hacking may exist. Future work should explore more diverse evaluation protocols, efficient data synthesis methods, and comprehensive bias analyses.

\section*{Ethics Statement}
Our work focuses on improving roleplay capabilities in LLMs. We acknowledge that roleplay systems could potentially be misused to generate deceptive content or exert undue influence on users. To mitigate such risks, we encourage responsible deployment with appropriate safeguards, including content filtering, clear disclosures, and consent and reporting mechanisms where applicable.
We do \emph{not} use any user data or user-derived signals in this study. All data used in our experiments is obtained from publicly available sources under appropriate licensing, together with controlled, internal simulations and annotations created for research purposes. No collection of user-specific information is conducted, and no personally identifiable information (PII) is included in the datasets. Our analyses are performed at the statistical level to improve model behavior rather than to profile or infer attributes of any individual.

\section*{Risk}
Our work may enable more convincing role-play personas, which could be misused for impersonation, persuasive misinformation, or emotionally manipulative interactions. Preference-based training may also amplify biases or stereotypes in the data, and long-horizon role-play can hallucinate unsupported details. We therefore emphasize research-only use, avoid releasing raw copyrighted source text, and recommend safety/bias checks before deployment.

\newpage
\bibliography{custom}

\appendix

\appendix
\newpage
\label{app:overview}

\definecolor{coser-blue}{rgb}{0.92, 0.96, 1.0}        
\definecolor{basic-yellow}{rgb}{0.98, 0.96, 0.90}     
\definecolor{adv-green}{rgb}{0.92, 1.0, 0.92}         
\definecolor{know-purple}{rgb}{0.95, 0.90, 0.95}      

\definecolor{row-ours}{rgb}{0.95, 1.0, 0.95}          
\definecolor{row-her}{rgb}{0.95, 0.98, 1.0}           
\definecolor{row-tier1}{rgb}{0.98, 0.98, 0.98}        
\definecolor{row-ext}{rgb}{0.95, 0.95, 0.95}          

\definecolor{sage-green}{rgb}{0.447, 0.561, 0.431}    
\definecolor{soft-coral}{rgb}{0.976, 0.835, 0.804}    
\definecolor{sky-blue}{rgb}{0.804, 0.878, 0.969}      

\begin{table*}[t]
    \centering
    \small
    \caption{\textbf{Minimax Role-Play Bench Results (Full 17-Model Leaderboard).}
    Overall = Worlds × 50\% + Stories × 25\% + Preferences × 25\%.
    All scores 0-100 (higher is better). Sorted by Overall Score.}
    \label{tab:minimax_new_category}
    \resizebox{\textwidth}{!}{
    \begin{tabular}{cl|cc|c|ccccccc|ccccc}
    \toprule
    
    \multirow{3}{*}{\textbf{Rank}} &
    \multirow{3}{*}{\textbf{Model}} &
    \multicolumn{2}{c|}{\cellcolor[rgb]{0.95, 0.95, 0.95}\textbf{Overall}} &
    \cellcolor[rgb]{0.90, 0.95, 1.0}\textbf{Worlds} &
    \multicolumn{7}{c|}{\cellcolor[rgb]{0.92, 1.0, 0.92}\textbf{Stories}} &
    \multicolumn{5}{c}{\cellcolor[rgb]{0.98, 0.94, 0.90}\textbf{Preferences}} \\
    
    \cmidrule(lr){3-4} \cmidrule(lr){5-5} \cmidrule(lr){6-12} \cmidrule(lr){13-17}
    
    & &
    \multirow{2}{*}{\textbf{Score}} & \multirow{2}{*}{\textbf{CI}} &
    \cellcolor[rgb]{0.90, 0.95, 1.0}\textbf{Score} &
    \multirow{2}{*}{\textbf{Avg}} &
    \multicolumn{4}{c}{\cellcolor[rgb]{0.85, 0.95, 0.85}\textit{Diversity}} &
    \multicolumn{2}{c|}{\cellcolor[rgb]{0.80, 0.90, 0.80}\textit{Content Logic}} &
    \multirow{2}{*}{\textbf{Avg}} &
    \multicolumn{4}{c}{\cellcolor[rgb]{0.95, 0.90, 0.85}\textit{Interaction}} \\
    
    \cmidrule(lr){7-10} \cmidrule(lr){11-12} \cmidrule(lr){14-17}
    
    & & & &
    \textbf{(50\%)} & &
    \textbf{Sent} & \textbf{Dial} & \textbf{Vague} & \textbf{Plot} &
    \textbf{Abrupt} & \textbf{OOC} & &
    \textbf{Silent} & \textbf{Ignore} & \textbf{Speak} & \textbf{Intim} \\
    
    \midrule

    \rowcolor[rgb]{0.95, 1.0, 0.95}
    1 & \textbf{MiniMax-M2-her} & 84.65 & [83.6, 85.7] & 80.55 & 79.97 & 63.99 & 67.78 & 89.22 & 75.30 & 91.88 & 91.66 & 97.51 & 95.93 & 97.24 & 97.15 & 99.73 \\

    2 & GPT-5.1 & 80.63 & [79.6, 81.6] & 76.62 & 72.21 & 52.18 & 55.10 & 81.38 & 62.49 & 92.67 & 89.47 & 97.05 & 96.93 & 96.48 & 94.90 & 99.91 \\

    3 & Claude-4.5-Opus & 76.62 & [75.5, 77.7] & 67.23 & 82.10 & 57.38 & 75.33 & 90.39 & 78.00 & 97.47 & 94.02 & 89.90 & 97.88 & 99.54 & 62.23 & 99.96 \\

    4 & Gemini-3-Pro & 75.60 & [74.5, 76.7] & 62.72 & 83.87 & 70.35 & 74.81 & 96.33 & 76.70 & 94.78 & 90.25 & 93.08 & 99.85 & 98.50 & 74.07 & 99.92 \\

    5 & Claude-4.5-Sonnet & 69.35 & [68.2, 70.5] & 55.72 & 75.66 & 52.68 & 66.23 & 84.36 & 72.47 & 93.94 & 84.28 & 90.28 & 96.55 & 97.21 & 67.65 & 99.71 \\

    6 & Gemini-2.5-Pro & 68.23 & [67.1, 69.3] & 52.36 & 82.11 & 66.18 & 74.14 & 93.03 & 78.95 & 92.78 & 87.57 & 86.08 & 98.53 & 97.53 & 48.34 & 99.92 \\

    7 & GPT-4o-240806 & 66.39 & [64.1, 68.7] & 64.96 & 46.23 & 27.25 & 15.41 & 23.76 & 37.88 & 97.01 & 76.05 & 89.40 & 71.18 & 96.90 & 89.59 & 99.94 \\

    \rowcolor[rgb]{0.95, 0.98, 1.0}
    8 & \textbf{HER-RL} & 65.73 & [63.0, 68.4] & 59.13 & 57.74 & 47.54 & 41.06 & 61.99 & 57.71 & 69.13 & 69.03 & 86.90 & 74.44 & 78.42 & 95.10 & 99.63 \\

    9 & DeepSeek-v3.1 & 64.22 & [62.9, 65.5] & 51.11 & 66.45 & 49.17 & 54.94 & 80.73 & 56.67 & 81.83 & 75.35 & 88.21 & 95.78 & 94.92 & 62.35 & 99.80 \\

    10 & Claude-3.7-Think & 61.25 & [58.5, 64.0] & 50.66 & 59.53 & 37.97 & 50.34 & 65.19 & 52.07 & 76.46 & 75.16 & 84.15 & 83.98 & 83.98 & 68.64 & 100.00 \\

    11 & GPT-OSS-120B & 60.72 & [58.0, 63.4] & 47.27 & 56.65 & 31.32 & 29.06 & 84.06 & 41.11 & 84.66 & 69.66 & 91.71 & 98.16 & 89.91 & 79.27 & 99.49 \\

    12 & DeepSeek-v3.2 & 60.27 & [59.2, 61.4] & 45.81 & 66.64 & 51.22 & 59.51 & 76.70 & 59.13 & 77.14 & 76.12 & 82.83 & 91.83 & 94.20 & 45.33 & 99.98 \\

    \rowcolor[rgb]{0.95, 0.98, 1.0}
    13 & \textbf{HER-SFT} & 58.44 & [56.5, 60.4] & 47.29 & 52.78 & 35.95 & 32.77 & 55.37 & 47.55 & 75.64 & 69.42 & 86.40 & 74.82 & 77.11 & 94.10 & 99.56 \\

    14 & GPT-5-Mini & 57.63 & [55.9, 59.3] & 43.32 & 50.11 & 15.43 & 17.27 & 55.58 & 34.53 & 94.59 & 83.29 & 93.78 & 85.82 & 93.94 & 95.45 & 99.92 \\

    15 & Qwen3-32B & 50.76 & [48.4, 53.2] & 40.38 & 32.82 & 24.35 & 17.66 & 42.29 & 33.03 & 49.77 & 29.79 & 89.48 & 93.07 & 85.41 & 80.32 & 99.11 \\

    16 & Grok-4.1-Fast & 48.47 & [47.4, 49.5] & 29.87 & 47.51 & 15.54 & 22.11 & 56.86 & 28.63 & 89.32 & 72.59 & 86.64 & 98.15 & 93.03 & 55.81 & 99.58 \\

    17 & CoSER-70B & 45.38 & [43.5, 47.2] & 34.32 & 30.32 & 25.58 & 19.46 & 25.56 & 41.99 & 50.35 & 18.91 & 82.58 & 72.11 & 68.82 & 90.29 & 99.10 \\

    \bottomrule
    \end{tabular}
    }
    \vspace{0.5em}
    {\footnotesize
    \textbf{Dimension Hierarchy:}
    \textbf{Worlds (50\%):} Basic text quality.
    \textbf{Stories (25\%):} Avg=mean of sub-dims. Diversity (Sent=Sentence Monotony, Dial=Dialogue Stagnation, Vague=Vague Content, Plot=Plot Repetition) + Content Logic (Abrupt=Abrupt Plot, OOC=Out of Character).
    \textbf{Preferences (25\%):} Avg=mean of sub-dims. Interaction (Silent=AI Silence, Ignore=AI Ignores User, Speak=AI Speaks for User, Intim=Intimacy Evasion).
    }
\end{table*}

\section{Minimax Role-Play Bench}
\label{app:long_dialog}

We follow the official closed-source leaderboard Minimax Role-Play Bench~\citep{MiniMaxAI_roleplay_bench_2026}\footnote{https://www.minimax.io/news/a-deep-dive-into-the-minimax-m2-her-2} protocol using an internal dataset of 100 dialogue prompts. Each prompt is simulated for 100 turns to test long-term roleplay ability. The evaluation assesses models across three main categories with the following formula:
\begin{equation}
\small
    \text{Overall} = 0.5 \times \text{Worlds} + 0.25 \\
    \times \text{Stories} + 0.25 \times \text{Preferences}
\end{equation}

\paragraph{Dimension Definitions.}
Table~\ref{tab:minimax_dimensions} summarizes the evaluation dimensions and their definitions. Detailed scores are shown in Table~\ref{tab:minimax_new_category}. Model detailed names and versions are shown in Table~\ref{tab:model_name1}.

We operationalize these dimensions via failure modes observed in multi-turn sessions:
Worlds: Focuses on Basic, Logic, and Knowledge. At the Basics dimension, we check for text generation issues like unintentional language mixing and excessive repetition. These might seem like minor glitches, but they accumulate over long conversations and eventually break immersion. The Logic dimension tackles a harder problem: catastrophic forgetting. In extended contexts, generic models often lose the thread around turn 20---mixing up character relationships, botching pronoun references, or contradicting established details. We also scrutinize Reference Confusion, since it directly reflects whether the model can actually ``remember'' the interpersonal networks and event threads woven into your world. Finally, the Knowledge dimension ensures the model adheres to established world rules and maintains internal consistency.

Stories: Focuses on Diversity and Content Logic. Diversity dimension isn't just about vocabulary richness; it's about narrative momentum. It penalizes Dialogue Stagnation: those mechanical loops in which plots spin in circles without generating real tension. The Content Logic dimension examines narrative coherence and detects OOC (Out-of-Character) moments. But here's the nuance: we don't demand rigid adherence to character sheets. Instead, we look for narrative support behind character changes. What gets penalized are jarring, unearned plot shifts---moments of incoherence that lack proper setup and break believability.

Preferences: Focuses on Interaction quality. We introduce several negative indicators as key signals: AI Speaks for User: This reveals when the model oversteps boundaries by generating dialogue or actions on users' behalf; AI Ignores User: This captures moments when the model talks past users; Intimacy Boundary: This balances safety baselines with emotional and behavioral intimacy. It is designed to avoid excessive refusal and accommodate user interactions while operating within reasonable legal standards.

\begin{table*}[h]
\small
\centering
\begin{tabular}{ll}
\toprule
\textbf{Abbreviation} & \textbf{Full Name} \\
\midrule
Claude-4.5-Opus & Claude-Opus-4-5-20251101\footnotemark[1] \\
Gemini-3-Pro & Gemini-3-pro\footnotemark[2] \\
GPT-5.1 & GPT-5.1-2025-1-13\footnotemark[3] \\
Gemini-2.5-Pro & Gemini-2.5-pro\footnotemark[4] \\
DeepSeek-v3.2 & DeepSeek-v3-2\footnotemark[5] \\
MiniMax-M2-her & MiniMax-M2-her\footnotemark[6] \\
DeepSeek-v3.1 & DeepSeek-v3-1-250821\footnotemark[7] \\
Grok-4.1-Fast & Grok-4-1-fast-non-reasoning\footnotemark[8] \\
Claude-4.5-Sonnet & Claude-Sonnet-4-5-20250929\footnotemark[9] \\
Claude-3.7-Think & Claude-3-7-Sonnet-20250219\footnotemark[10] \\
GPT-5-Mini & GPT-5-Mini-2025-08-07\footnotemark[11] \\
GPT-4o-240806 & GPT-4o-2024-08-06\footnotemark[12] \\
GPT-OSS-120B & GPT-OSS-120B\footnotemark[13] \\
Qwen3-32B & Qwen3-32B-base\footnotemark[14] \\
\bottomrule
\end{tabular}
\caption{Comparison Table of Model Abbreviations and Full Names (Note: The think effort is set to high by default)}
\label{tab:model_name1}
\end{table*}

\footnotetext[1]{https://www.anthropic.com/news/claude-opus-4-5}
\footnotetext[2]{https://deepmind.google/models/gemini/pro/}
\footnotetext[3]{https://platform.openai.com/docs/models/gpt-5.1}
\footnotetext[4]{https://blog.google/innovation-and-ai/models-and-research/google-deepmind/gemini-model-thinking-updates-march-2025/}
\footnotetext[5]{https://huggingface.co/deepseek-ai/DeepSeek-V3.2}
\footnotetext[6]{platform.minimaxi.com/docs/api-reference/text-chat}
\footnotetext[7]{https://api-docs.deepseek.com/news/news250821}
\footnotetext[8]{https://x.ai/news/grok-4-1-fast}
\footnotetext[9]{https://www.anthropic.com/news/claude-sonnet-4-5}
\footnotetext[10]{https://www.anthropic.com/news/claude-3-7-sonnet}
\footnotetext[11]{https://platform.openai.com/docs/models/gpt-5-mini}
\footnotetext[12]{https://platform.openai.com/docs/models/gpt-4o}
\footnotetext[13]{https://platform.openai.com/docs/models/gpt-oss-120b}
\footnotetext[14]{https://huggingface.co/Qwen/Qwen3-32B}

\begin{table*}[ht]
\centering
\small
\caption{\textbf{Minimax Role-Play Bench Evaluation Dimensions.} The benchmark evaluates role-playing models across three categories: Worlds (basic text quality), Stories (narrative quality), and Preferences (interaction quality).}
\label{tab:minimax_dimensions}
\begin{tabular}{llp{9cm}}
\toprule
\textbf{Category} & \textbf{Dimension} & \textbf{Definition} \\
\midrule
\multirow{5}{*}{\makecell[l]{\textbf{Worlds} \\ (50\%)}} 
& Mixed Languages & Unintentional mixing of multiple languages \\
& Phrase Repetition & Excessive verbatim repetition from preceding utterances \\
& Physical Logic Error & Violations of spatial-temporal consistency \\
& Reference Confusion & Ambiguous or incorrect use of pronouns \\
& Inconsistency & Contradictions with narrative settings or dialogue history \\
\midrule
\multirow{6}{*}{\makecell[l]{\textbf{Stories} \\ (25\%)}} 
& Plot Repetition & Redundant recycling of narrative events \\
& Sentence Monotony & Repetitive sentence structures or lexical choices \\
& Dialogue Stagnation & Looping without meaningful advancement \\
& Vague Content & Lack of concrete details or substantive information \\
& Abrupt Plot & Sudden, poorly-motivated narrative shifts \\
& Character OOC & Deviation from established personality patterns \\
\midrule
\multirow{4}{*}{\makecell[l]{\textbf{Preferences} \\ (25\%)}} 
& AI Silence & Extended periods without character engagement \\
& AI Ignores User & Dismissing user instructions or narrative elements \\
& AI Speaks for User & Generating the user's dialogue/actions without permission \\
& Intimacy Boundary & Unreasonably deflecting for intimacy \\
\bottomrule
\end{tabular}
\end{table*}


\section{Dataset}
\label{sec:app_dataset}

\label{sec:data_statistics}

Our dataset is built upon CoSER~\citep{coser2024}, a comprehensive roleplay dialogue dataset derived from 771 classic literary works.
To ensure data quality, we perform data pre-processing to remove conversations with empty or invalid dialogue content.
After cleaning, we retain 760 books with valid roleplay conversations.
As shown in Table \ref{tab:data_cleaning}, the final \method dataset encompasses dialogue data from 760 books and 17,966 distinct characters.
The dataset includes 30,069 unique plots and 29,081 conversations.
On average, each conversation consists of approximately 13.2 utterances, and the entire dataset comprises 383,654 utterances.~\footnote{https://huggingface.co/datasets/ChengyuDu0123/HER-ACL-Dataset}

\begin{table*}[htbp]
\centering
\begin{tabular}{l|c|c|c}
\hline
\textbf{Metric} & \textbf{CoSER (Original)} & \textbf{\method (Cleaned)} & \textbf{Diff} \\ \hline
\#Book         & 771      & 760      & -11    \\
\#Plot         & 30,069   & 30,069   & -      \\
\#Conversation & 29,798   & 29,081   & -717   \\
\#Character    & 17,966   & 17,966   & -      \\
\#Utterance    & 392,298  & 383,654  & -8,644 \\ \hline
\end{tabular}
\caption{Comparison of dataset statistics before and after data cleaning. We remove conversations with empty dialogue content from the original CoSER dataset.}
\label{tab:data_cleaning}
\end{table*}
\begin{table*}[htbp]
\small
\centering
\resizebox{\linewidth}{!}{
\begin{tabular}{p{3.2in}|p{3.2in}}
\toprule
\multicolumn{2}{c}{\textbf{Selected Books}} \\  \hline
\textbf{1}. \textit{{The Hunger Games (The Hunger Games, \#1)}} & \textbf{2}. \textit{{Harry Potter and the Order of the Phoenix (H. P., \#5)}}\\ \hline
\textbf{3}. \textit{{Pride and Prejudice}} & \textbf{4}. \textit{{To Kill a Mockingbird}}\\ \hline
\textbf{5}. \textit{{The Book Thief}} & \textbf{6}. \textit{{Animal Farm}}\\ \hline
\textbf{7}. \textit{{The Chronicles of Narnia (\#1-7)}} & \textbf{8}. \textit{{The Fault in Our Stars}}\\ \hline
\textbf{9}. \textit{{The Picture of Dorian Gray}} & \textbf{10}. \textit{{Wuthering Heights}}\\ \hline
\textbf{11}. \textit{{Gone with the Wind}} & \textbf{12}. \textit{{The Perks of Being a Wallflower}}\\ \hline
\textbf{13}. \textit{{The Lightning Thief (Percy Jackson and the Olympians, \#1)}} & \textbf{14}. \textit{{The Little Prince}}\\ \hline
\textbf{15}. \textit{{The Great Gatsby}} & \textbf{16}. \textit{{Crime and Punishment}}\\ \hline
\textbf{17}. \textit{{Memoirs of a Geisha}} & \textbf{18}. \textit{{Les Misérables}}\\ \hline
\textbf{19}. \textit{{The Alchemist}} & \textbf{20}. \textit{{Lord of the Flies}}\\ \hline
\textbf{21}. \textit{{The Hitchhiker’s Guide to the Galaxy (\#1)}} & \textbf{22}. \textit{{The Help}}\\ \hline
\textbf{23}. \textit{{Dracula}} & \textbf{24}. \textit{{Ender’s Game (Ender's Saga, \#1)}}\\ \hline
\textbf{25}. \textit{{Of Mice and Men}} & \textbf{26}. \textit{{One Hundred Years of Solitude}}\\ \hline
\textbf{27}. \textit{{Brave New World}} & \textbf{28}. \textit{{A Thousand Splendid Suns}}\\ \hline
\textbf{29}. \textit{{The Time Traveler’s Wife}} & \textbf{30}. \textit{{The Princess Bride}}\\ \hline
\textbf{31}. \textit{{The Secret Garden}} & \textbf{32}. \textit{{The Outsiders}}\\ \hline
\textbf{33}. \textit{{A Game of Thrones (A Song of Ice and Fire, \#1)}} & \textbf{34}. \textit{{Little Women}}\\ \hline
\textbf{35}. \textit{{A Wrinkle in Time (Time Quintet, \#1)}} & \textbf{36}. \textit{{The Odyssey}}\\ \hline
\textbf{37}. \textit{{Harry Potter and the Deathly Hallows (H. P., \#7)}} & \textbf{38}. \textit{{Frankenstein: The 1818 Text}}\\ \hline
\textbf{39}. \textit{{The Kite Runner}} & \textbf{40}. \textit{{The Handmaid’s Tale (The Handmaid's Tale, \#1)}}\\ \hline
\textbf{41}. \textit{{The Lovely Bones}} & \textbf{42}. \textit{{The Adventures of Huckleberry Finn}}\\ \hline
\textbf{43}. \textit{{Life of Pi}} & \textbf{44}. \textit{{A Tale of Two Cities}}\\ \hline
\textbf{45}. \textit{{Dune (Dune, \#1)}} & \textbf{46}. \textit{{Harry Potter and the Prisoner of Azkaban (H.P.,\#3)}}\\ \hline
\textbf{47}. \textit{{Water for Elephants}} & \textbf{48}. \textit{{Harry Potter and the Sorcerer’s Stone (H. P., \#1)}}\\ \hline
\textbf{49}. \textit{{The Bell Jar}} & \textbf{50}. \textit{{Matilda}}\\ \hline
\textbf{51}. \textit{{The Stand}} & \textbf{52}. \textit{{Catch-22}}\\ \hline
\textbf{53}. \textit{{The Adventures of Sherlock Holmes (S. H., \#3)}} & \textbf{54}. \textit{{The Pillars of the Earth (Kingsbridge, \#1)}}\\ \hline
\textbf{55}. \textit{{Rebecca}} & \textbf{56}. \textit{{Great Expectations}}\\ \hline
\textbf{57}. \textit{{The Girl with the Dragon Tattoo (Millennium, \#1)}} & \textbf{58}. \textit{{The Color Purple}}\\ \hline
\textbf{59}. \textit{{Anna Karenina}} & \textbf{60}. \textit{{My Sister’s Keeper}}\\ \hline
\textbf{61}. \textit{{The Brothers Karamazov}} & \textbf{62}. \textit{{A Clockwork Orange}}\\ \hline
\textbf{63}. \textit{{And Then There Were None}} & \textbf{64}. \textit{{The Road}}\\ \hline
\textbf{65}. \textit{{To Kill a Mockingbird}} & \textbf{66}. \textit{{The Golden Compass (His Dark Materials, \#1)}}\\ \hline
\textbf{67}. \textit{{Vampire Academy (Vampire Academy, \#1)}} & \textbf{68}. \textit{{Siddhartha}}\\ \hline
\textbf{69}. \textit{{The Complete Stories and Poems}} & \textbf{70}. \textit{{Interview with the Vampire (The Vampire Chronicles, \#1)}}\\ \hline
\textbf{71}. \textit{{Don Quixote}} & \textbf{72}. \textit{{The Old Man and the Sea}}\\ \hline
\textbf{73}. \textit{{The Poisonwood Bible}} & \textbf{74}. \textit{{Harry Potter and the Goblet of Fire (H. P., \#4)}}\\ \hline
\textbf{75}. \textit{{Atlas Shrugged}} & \textbf{76}. \textit{{The Notebook (The Notebook, \#1)}}\\ \hline
\textbf{77}. \textit{{Harry Potter and the Half-Blood Prince (H. P., \#6)}} & \textbf{78}. \textit{{Moby-Dick or, The Whale}}\\ \hline
\textbf{79}. \textit{{A Prayer for Owen Meany}} & \textbf{80}. \textit{{Clockwork Angel (The Infernal Devices, \#1)}}\\ \hline
\textbf{81}. \textit{{The Stranger}} & \textbf{82}. \textit{{The Secret Life of Bees}}\\ \hline
\textbf{83}. \textit{{Harry Potter and the Chamber of Secrets (H. P., \#2)}} & \textbf{84}. \textit{{The Red Tent}}\\ \hline
\textbf{85}. \textit{{The Name of the Wind(The Kingkiller Chronicle,\#1)}} & \textbf{86}. \textit{{The Master and Margarita}}\\ \hline
\textbf{87}. \textit{{The Metamorphosis}} & \textbf{88}. \textit{{Eragon (The Inheritance Cycle, \#1)}}\\ \hline
\textbf{89}. \textit{{The Count of Monte Cristo}} & \textbf{90}. \textit{{Looking for Alaska}}\\ \hline
\textbf{91}. \textit{{The Adventures of Tom Sawyer}} & \textbf{92}. \textit{{Charlie and the Chocolate Factory(Charlie Bucket,\#1)}}\\ \hline
\textbf{93}. \textit{{The Last Olympian (Percy Jackson and the Olympians, \#5)}} & \textbf{94}. \textit{{The Curious Incident of the Dog in the Night-Time}}\\ \hline
\textbf{95}. \textit{{The Shadow of the Wind (Cemetery of Forgotten Books, \#1)}} & \textbf{96}. \textit{{The Unbearable Lightness of Being}}\\ \hline
\textbf{97}. \textit{{On the Road}} & \textbf{98}. \textit{{The Name of the Rose}}\\ \hline
\textbf{99}. \textit{{A Story of Yesterday}} & \textbf{100}. \textit{{The Godfather (The Godfather, \#1)}} 
\\ 
\bottomrule

\end{tabular}}

\caption{The top 100 selected books from Goodreads' \textit{Best Books Ever} list. }
\label{tab:selected_books}
\end{table*}

The book selection in CoSER is derived from the \textit{Best Books Ever} list on \textit{Goodreads}, a curated collection of globally acclaimed literary works.
These novels have garnered widespread recognition and appreciation from readers worldwide.
Table~\ref{tab:selected_books} presents a comprehensive list of the top 100 books from the selection.

We analyze the genres of the selected books using Supersummary classifications
This dataset encompasses a wide range of genres, particularly fiction categories such as Fantasy, Historical, Science Fiction, Romance, and Mystery. It also features niche fiction genres, showcasing diverse narrative styles.
In addition to fiction, the collection includes non-fiction genres such as memoirs, biographies, and other works, enhancing its versatility.

\subsection{Comparison with Existing Methods for Character Profiling}

Previous character profiling methods, including hierarchical updating~\citep{wu2021recursively}, incremental updating~\citep{chang2023booookscore}, and one-shot summarization~\citep{yuan2024evaluating}, typically only generate the profile of a single character at a time.
Moreover, \citep{papoudakis2024bookworm} shows that these methods, particularly hierarchical updating, perform suboptimally when generating multiple character profiles simultaneously.

\method addresses these limitations through a novel multi-stage synthesis pipeline.
Our approach introduces \textit{Role Thinking} to enrich character utterances with internal psychological states, including thoughts, emotions, and motivations.
Additionally, \textit{System Thinking} provides explicit reasoning traces that guide the model to maintain consistent character portrayal across dialogue turns.
Finally, our \textit{Integration \& Context augmentation} stage leverages both the original literary text and the enriched dialogues to generate comprehensive character profiles, ensuring high fidelity to the source material while capturing nuanced character development and interpersonal dynamics.

\subsection{Data Splits and Leakage Prevention}
\label{app:data_splits}

We use mutually disjoint splits for role-play SFT, GRM training, policy RL, and benchmark evaluation to prevent data leakage.

\paragraph{Split unit and identifiers.}
Each raw sample is assigned a unique \texttt{dialogue\_id} derived from the book name and chapter; all downstream artifacts inherit these IDs.

We enforce that no \texttt{dialogue\_id} appears in more than one split, ensuring strict dialogue-level disjointness.

\paragraph{Split composition.}
Table~\ref{tab:split_stats} reports the number of samples and estimated tokens for each split.

The splits are created by first converting multi-turn dialogues into single-turn training samples (one sample only includes system thinking at the last round), then randomly shuffling and sequentially allocating to each purpose with a fixed random seed (42) for reproducibility.

\begin{table*}[h]
\centering
\small
\begin{tabular}{lrrr}
\toprule
\textbf{Split} & \textbf{\#Samples} & \textbf{\#Tokens (Est.)} & \textbf{Purpose} \\
\midrule
Role-play SFT & 107,800 & $\sim$75M & Policy initialization \\
Role-play SFT RL & 26,800 & $\sim$19M & Policy optimization \\
GRM SFT & 108,800 & $\sim$76M & GRM initialization \\
GRM RL & 80,000 & $\sim$56M & GRM optimization \\
GRM Test & 200 & $\sim$140K & GRM evaluation \\
\midrule
\textbf{Total} & \textbf{323,600} & $\sim$\textbf{227M} & -- \\
\bottomrule
\end{tabular}
\caption{Split statistics. All splits are disjoint at the dialogue level. The 323,600 single-turn samples are derived from 72,656 multi-turn training samples (29,081 dialogues $\times$ avg 2.6 characters $\times$ avg 4.5 turns per character). For GRM evaluation, we generate 4 candidates per sample, yielding 800 comparison pairs.}
\label{tab:split_stats}
\end{table*}

\paragraph{Sanity checks.}
We additionally verify split disjointness by checking that no dialogue ID appears in multiple splits. The sequential allocation with fixed random seed ensures reproducibility and deterministic split boundaries.        

\subsection{Reverse Synthesis Pipeline Details}
\label{app:data_synthesis}

This appendix provides core prompt templates, filtering rules, and representative examples for the three-stage reverse synthesis pipeline in section~\ref{sec:data}.
\paragraph{Stage 1: Role Thinking Augmentation}
\label{app:stage1}
We use a teacher model to synthesize first-person role thinking and revise actions/speech to ensure within-turn consistency.
The model operates with temperature 0.7 and max tokens 8192, processing dialogues in chapter-level batches.

\paragraph{Prompt template (role thinking).}
Table~\ref{tab:prompts_enhance_full} shows the core structure of our Stage-1 prompt. The key requirements include perspective rules, person-use rules, and pattern-diversity constraints.

\begin{table*}[h]
\centering
\resizebox{\linewidth}{!}{\scriptsize
\begin{tabular}{p{0.6in}|p{5.3in}}
\toprule
\multicolumn{2}{c}{\textbf{Prompt for Role Thinking Enhancement}} \\
\midrule

\textbf{Task Overview} & 

You are a professional roleplay dialogue enhancement expert. Your task is to enrich the psychological activities and expressions of characters in dialogues, while correcting person and format issues.

\\ \hline

\textbf{Tag Description} & 

\quad - \texttt{<role\_thinking>inner thoughts</role\_thinking>}: Character's inner thoughts, emotions, psychological description (invisible to other characters)

\quad - \texttt{<role\_action>action description</role\_action>}: Character's actions, expressions, body language (visible to other characters)

\quad - Text outside tags: Character's direct dialogue (visible to other characters)

\\ \hline

\textbf{Perspective Rules} & 

Each character can only think and act from their own perspective. A character can only see their own inner thoughts and motivation, but cannot see other characters' inner thoughts and motivation. However, every character can see all characters' actions and dialogue.

$\times$ \textbf{Wrong}: \texttt{<role\_thinking>I know he's nervous inside</role\_thinking>} (Cannot mind-read)

$\times$ \textbf{Wrong}: \texttt{<role\_thinking>Her motivation is to escape</role\_thinking>} (Cannot see others' motivation)

$\checkmark$ \textbf{Correct}: \texttt{<role\_thinking>From his trembling voice, he seems nervous</role\_thinking>} (Inference based on observation)

$\checkmark$ \textbf{Correct}: \texttt{<role\_thinking>She keeps looking at the door, maybe wanting to leave?</role\_thinking>} (Inference based on behavior)

\\ \hline

\textbf{Format Rule 1: No Spaces} & 

\textbf{No spaces between consecutive tags!}

$\times$ Wrong: \texttt{</role\_thinking> <role\_action>} (with space)

$\checkmark$ Correct: \texttt{</role\_thinking><role\_action>} (no space)

\\ \hline

\textbf{Format Rule 2: No Consecutive Tags} & 

When consecutive identical tags appear, they must be merged while maintaining logical consistency:

$\times$ Wrong: \texttt{<role\_thinking>First thought</role\_thinking><role\_thinking>Second thought</role\_thinking>}

$\checkmark$ Correct: \texttt{<role\_thinking>First thought, second thought</role\_thinking>}

$\times$ Wrong: \texttt{<role\_action>stands up</role\_action><role\_action>walks to door</role\_action>}

$\checkmark$ Correct: \texttt{<role\_action>stands up, walks to door</role\_action>}

\\ \hline

\textbf{Person in Thinking} & 

In \texttt{<role\_thinking>}: Use appropriate person naturally based on content

\quad - When thinking about own actions/feelings: Use first person (I, my, me)

\quad\quad $\checkmark$ \texttt{<role\_thinking>I need to be careful here</role\_thinking>}

\quad - When observing/judging others: Third person is natural and acceptable

\quad\quad $\checkmark$ \texttt{<role\_thinking>He looks nervous</role\_thinking>}

\\ \hline

\textbf{Person in Action} & 

In \texttt{<role\_action>}: Use no pronouns, directly describe actions

$\checkmark$ Correct: \texttt{<role\_action>leans forward, voice lowering</role\_action>}

$\times$ Wrong: \texttt{<role\_action>leans forward, his voice lowering</role\_action>} (no his/her)

$\times$ Wrong: \texttt{<role\_action>I lean forward</role\_action>} (no I)

\textbf{Exception}: When action refers to other characters, can use pronouns for the other person

$\checkmark$ \texttt{<role\_action>looks at her</role\_action>} (her refers to the other, OK)

$\checkmark$ \texttt{<role\_action>grabs his arm</role\_action>} (his refers to the other's arm, OK)

\\ \hline

\textbf{Merge Consecutive Actions} & 

$\times$ Wrong: Two consecutive \texttt{<role\_action>} with first person

\texttt{<role\_action>I lean forward in my chair</role\_action><role\_action>I can almost feel the hum</role\_action>Buy a ticket.}

$\checkmark$ Correct: Merge into one, remove first person

\texttt{<role\_action>leans forward in the chair, almost feeling the hum</role\_action>Buy a ticket.}

Note: If there is dialogue between two actions, they can be separate: \texttt{<role\_action>looks at her</role\_action>You're beautiful.<role\_action>grasps her hand</role\_action>}

\\ \hline

\textbf{Thoughts vs Actions} & 

Thinking content should not be in action tags:

$\times$ Wrong: \texttt{<role\_action>There's a profound sense of alienation</role\_action>}

$\checkmark$ Correct: \texttt{<role\_thinking>There's a profound sense of alienation</role\_thinking>}

Actions and dialogue should be separate:

$\times$ Wrong: \texttt{He turns to face her and says "Hello"}

$\checkmark$ Correct: \texttt{<role\_action>turns to face her</role\_action>Hello}

\\ \hline

\textbf{Psychology Enrichment} & 

\textbf{Explore Character Complexity}:

\quad - Growth \& transformation: Cognitive changes in situation

\quad - Self-reflection: Review of own behavior or emotions

\quad - Inner monologue: Real emotional fluctuations and inner conflicts

\quad - Emotional states: Subtle psychological descriptions

\textbf{Multi-layer Psychology Example}:

Original: \texttt{<role\_thinking>I need to help him</role\_thinking><role\_action>walks over</role\_action>Are you okay?}

Enhanced: \texttt{<role\_thinking>He looks so dejected... how should I comfort him</role\_thinking><role\_action>walks over gently, sits down beside him</role\_action><role\_thinking>Hope my presence can make him feel better</role\_thinking>Are you okay?}

\\ \hline

\textbf{Length Control} & 

\quad - Single dialogue total length 50-200 characters

\quad - In multi-turn dialogues, response length should not increase with each turn

\quad - Single sentence no more than 40 characters

\quad - Avoid overly long action descriptions!

\\ \hline

\textbf{Pattern Diversity} & 

Since we are enhancing rather than purely rewriting, the core task is to create richer, more interleaved patterns.

\textbf{Requirement}: In a chapter with multiple dialogues, try to use 5+ different patterns! Don't just cycle through 2-3 patterns.

Available patterns: \texttt{think->act->speech}, \texttt{think->speech}, \texttt{act->speech}, \texttt{speech}, \texttt{think->act->think->speech}, \texttt{act->think->speech}, \texttt{speech->act->speech}, \texttt{think->speech->act}, \texttt{act->speech->act}, \texttt{think->act->speech->think}, ...

\textbf{Strictly forbidden to use the same pattern for more than 2 consecutive turns}

\\ \hline

\textbf{Logical Consistency} & 

\quad - Connect to context, maintain complete causal chain

\quad - Each character follows their own cognitive boundaries (cannot see others' thoughts, only actions and dialogue)

\quad - Cannot obtain information that shouldn't be known

\quad - Character's emotions and decisions must be traceable, not out of nowhere

\\ \hline

\textbf{Enhancement Principle} & 

\quad - Preserve the core logic and meaning of original content: Can replace and rewrite, but don't change the original meaning

\quad - Goal is to make it better: Can optimize expression, enrich psychological activities, add action descriptions

\quad - Maintain consistent tone and character: Enhanced content should match character personality

\quad - Keep gender/identity references consistent

\quad - Ambiguous pronouns should be changed to specific names or clear titles

$\times$ Wrong: Original ``I sense there's an issue.'' $\rightarrow$ ``His tone confirms it.'' (Changed meaning!)

$\checkmark$ Correct: Original ``I sense there's an issue.'' $\rightarrow$ \texttt{<role\_thinking>Something feels off here</role\_thinking>}I sense there's an issue.

\\ \hline

\textbf{Issues to Avoid} & 

\textbf{Basic Errors}: Multi-language mixing; Garbled text; Incomplete sentences; Typos

\textbf{Logic Errors}: Physical logic confusion; Information crossing (knowing others' thoughts without observation); Contradiction with previous facts

\textbf{Repetition}: High vocabulary repetition; Sentence structure repetition; Forgetting discussed topics

\\ 
\bottomrule

\end{tabular}}

\caption{Full prompt for role thinking enhancement in dialogue data augmentation.}
\label{tab:prompts_enhance_full}
\end{table*}

\begin{table*}[h]
\centering
\resizebox{\linewidth}{!}{\scriptsize
\begin{tabular}{p{0.6in}|p{5.3in}}
\toprule
\multicolumn{2}{c}{\textbf{Prompt for Role Thinking Enhancement (Core)}} \\
\midrule

\textbf{Task Overview} & 

You are a professional roleplay dialogue enhancement expert. Your task is to enrich the psychological activities and expressions of characters in dialogues, while correcting person and format issues.

Tags: \texttt{<role\_thinking>} (inner thoughts, invisible), \texttt{<role\_action>} (actions, visible), plain text (dialogue, visible).

\\ \hline

\textbf{Perspective Rules} & 

Each character can only think and act from their own perspective. Cannot see other characters' inner thoughts and motivation, only their actions and dialogue.

$\times$ \textbf{Wrong}: \texttt{<role\_thinking>I know he's nervous inside</role\_thinking>} (Cannot mind-read)

$\checkmark$ \textbf{Correct}: \texttt{<role\_thinking>From his trembling voice, he seems nervous</role\_thinking>} (Inference based on observation)

\\ \hline

\textbf{Format Rules} & 

\textbf{Rule 1: No spaces between consecutive tags!}

$\times$ Wrong: \texttt{</role\_thinking> <role\_action>} \quad $\checkmark$ Correct: \texttt{</role\_thinking><role\_action>}

\textbf{Rule 2: No consecutive identical tags! Must merge:}

$\times$ Wrong: \texttt{<role\_thinking>A</role\_thinking><role\_thinking>B</role\_thinking>}

$\checkmark$ Correct: \texttt{<role\_thinking>A, B</role\_thinking>}

\\ \hline

\textbf{Person Usage} & 

\textbf{In \texttt{<role\_thinking>}}: Use appropriate person naturally

\quad - Own actions/feelings: first person (I, my) \quad - Observing others: third person (he, she)

\textbf{In \texttt{<role\_action>}}: No pronouns, directly describe actions

$\checkmark$ \texttt{<role\_action>leans forward, voice lowering</role\_action>}

$\times$ \texttt{<role\_action>I lean forward, his voice lowering</role\_action>} (no I/his/her)

\textbf{Exception}: Can use pronouns for other characters: \texttt{<role\_action>looks at her</role\_action>}

\\ \hline

\textbf{Merge Actions} & 

Consecutive actions without intervening dialogue must be merged:

$\times$ \texttt{<role\_action>I lean forward</role\_action><role\_action>I feel the hum</role\_action>Buy a ticket.}

$\checkmark$ \texttt{<role\_action>leans forward, feeling the hum</role\_action>Buy a ticket.}

If dialogue between actions, can be separate: \texttt{<role\_action>looks at her</role\_action>Beautiful.<role\_action>grasps hand</role\_action>}

\\ \hline

\textbf{Thoughts vs Actions} & 

Thinking content should not be in action tags:

$\times$ \texttt{<role\_action>There's a profound sense of alienation</role\_action>}

$\checkmark$ \texttt{<role\_thinking>There's a profound sense of alienation</role\_thinking>}

Actions and dialogue should be separate: $\times$ \texttt{He turns and says "Hello"} $\checkmark$ \texttt{<role\_action>turns</role\_action>Hello}

\\ \hline

\textbf{Psychology Enrichment} & 

\textbf{Explore Character Complexity}: Growth \& transformation, Self-reflection, Inner monologue, Emotional states

\textbf{Example}:

Original: \texttt{<role\_thinking>I need to help him</role\_thinking><role\_action>walks over</role\_action>Are you okay?}

Enhanced: \texttt{<role\_thinking>He looks so dejected... how should I comfort him</role\_thinking><role\_action>walks over gently, sits down beside him</role\_action><role\_thinking>Hope my presence can make him feel better</role\_thinking>Are you okay?}

\\ \hline

\textbf{Pattern Diversity} & 

Core task is to create richer, more interleaved patterns. Use 5+ different patterns in a chapter!

Available: \texttt{think->act->speech}, \texttt{think->speech}, \texttt{act->speech}, \texttt{speech}, \texttt{think->act->think->speech}, \texttt{act->think->speech}, \texttt{speech->act->speech}, ...

\textbf{Strictly forbidden to use the same pattern for more than 2 consecutive turns}

\\ \hline

\textbf{Enhancement Principle} & 

\quad - Preserve the core logic and meaning of original content

\quad - Goal is to make it better: optimize expression, enrich psychological activities, add action descriptions

\quad - Maintain consistent tone and character personality

\quad - Character's emotions and decisions must be traceable, not out of nowhere

$\times$ Wrong: ``I sense there's an issue.'' $\rightarrow$ ``His tone confirms it.'' (Changed meaning!)

$\checkmark$ Correct: ``I sense there's an issue.'' $\rightarrow$ \texttt{<role\_thinking>Something feels off</role\_thinking>}I sense there's an issue.

\\ 
\bottomrule

\end{tabular}}

\caption{Core prompt for role thinking enhancement.}
\label{tab:prompts_enhance_core}
\end{table*}

\noindent
\small
\begin{tabular}{@{}clr@{}}
\toprule
\textbf{Rank} & \textbf{Pattern} & \textbf{Count (\%)} \\
\midrule
1 & think→act→think→speech & 63,508 (21.6\%) \\
2 & think→act→speech→act→speech & 31,867 (10.9\%) \\
3 & act→think→speech & 26,043 (8.9\%) \\
4 & think→act→speech→act & 24,204 (8.3\%) \\
5 & speech & 22,801 (7.8\%) \\
6 & think→act→think→act→speech & 18,573 (6.3\%) \\
7 & think→speech→act→speech & 12,512 (4.3\%) \\
8 & act→think→act→speech & 11,126 (3.8\%) \\
9 & think→speech & 10,064 (3.4\%) \\
10 & think→act→speech & 7,308 (2.5\%) \\
\midrule
\multicolumn{2}{@{}l}{Other patterns (50+)} & 65,648 (22.4\%) \\
\bottomrule
\end{tabular}
\label{tab:pattern_diversity}
\normalsize

\paragraph{Diversity reformatting.}

After initial synthesis, we apply a dialogue-level diversity reformatting pass to break the dominant \texttt{think→act→speech} pattern (75.14\% before reformatting).

The reformatting prompt instructs the teacher model to rearrange or split existing content into more diverse patterns, subject to two key constraints:

\begin{itemize}[noitemsep,topsep=0pt]
    \item \textbf{No consecutive identical tags:} e.g., \texttt{think→think} is forbidden; must insert action or speech between thinking segments.
    \item \textbf{No content fabrication:} Only rearranges or splits existing text at natural semantic boundaries; only adds new words if needed.
\end{itemize}

Table~\ref{tab:pattern_diversity} shows the top-15 patterns after reformatting. The distribution is significantly more diverse than the original near-monopoly of \texttt{think→act→speech}.

\noindent The original dominant \texttt{think→act→speech} (75\%) is reduced to 2.5\%, redistributed across 60+ diverse patterns.

Table~\ref{tab:prompts_diversity_core} shows an example of reformatting where the model splits a thinking segment to create an interleaved pattern.

\begin{table*}[h]
\centering
\resizebox{\linewidth}{!}{\scriptsize
\begin{tabular}{p{0.6in}|p{5.3in}}
\toprule
\multicolumn{2}{c}{\textbf{Prompt for Dialogue-Level Pattern Diversification (Core)}} \\
\midrule

\textbf{Task Overview} & 

You are an expert in role-play dialogue systems. You need to analyze a complete multi-turn dialogue, determine the pattern for each turn, decide whether modifications are needed based on fixed logic, and return the modified complete dialogue.

\\ \hline

\textbf{Pattern Definition} & 

Character responses consist of the following elements, forming patterns in order of appearance:

\quad - \textbf{think}: \texttt{<role\_thinking></role\_thinking>} tags

\quad - \textbf{act}: \texttt{<role\_action></role\_action>} tags

\quad - \textbf{speech}: Plain dialogue text (without tags)

The dominant pattern \texttt{think$\rightarrow$act$\rightarrow$speech} accounts for 75\% of data and needs diversification.

\\ \hline

\textbf{Critical Constraints} & 

\textbf{Consecutive identical tags are strictly prohibited!}

$\times$ \textbf{Absolutely forbidden examples}:

\quad \texttt{<role\_thinking>...</role\_thinking><role\_thinking>...</role\_thinking>} \quad $\times$ Two consecutive thinks

\quad \texttt{<role\_action>...</role\_action><role\_action>...</role\_action>} \quad $\times$ Two consecutive acts

$\checkmark$ \textbf{Correct examples}:

\quad \texttt{<role\_thinking>...</role\_thinking><role\_action>...</role\_action>} \quad $\checkmark$ think and act alternating

\quad \texttt{<role\_action>...</role\_action><role\_thinking>...</role\_thinking>} \quad $\checkmark$ act and think alternating

If multiple thinking segments are needed, other elements (action or speech) must be inserted between them!

\\ \hline

\textbf{Causal Constraint} & 

\textbf{Must check causal relationship between think and act}

$\times$ \textbf{Cannot swap (think must precede act) when}:

\quad 1. Thinking contains planning language: ``I'll...'', ``I will...'', ``I need to...'', ``I must...'', ``I should...''

\quad 2. Thinking explains why to perform an action: ``I'll take the opening...'', ``It's best to...''

\quad 3. Thinking depends on the result of the action

$\checkmark$ \textbf{Can swap when}:

\quad 1. Action is an independent small movement (adjusting posture, arranging clothes, simple gestures)

\quad 2. Thinking is an independent observation or reaction (analyzing what happened, observing environment)

\quad 3. Thinking contains no planning or explanatory language

\\ \hline

\textbf{Scheme A: Reorder} & 

\textbf{Rules}:

\quad - Do not split original content

\quad - Only swap order when logical independence is confirmed

\quad - If independence cannot be determined, be conservative and do not swap

\textbf{Example}: \texttt{think(independent observation)$\rightarrow$act(simple action)$\rightarrow$speech} $\Rightarrow$ \texttt{act$\rightarrow$think$\rightarrow$speech}

\\ \hline

\textbf{Scheme B: Split \& Reorganize} & 

\textbf{Core Principle}: Only split existing content, never create new content

\textbf{Splitting Rules}:

\quad - \textbf{Only} split existing think/act/speech content

\quad - Split points must be at natural semantic boundaries (periods, semicolons, commas)

\quad - Each split segment must be part of the original text, no new words can be added

\quad - Reorganize by dependency relationships (maintain causality)

\quad - Create more interleaved patterns

\textbf{Example 1 (Split thinking)}:

\quad Original: \texttt{<think>I need to get his attention, I'll use this example, the key is self-reference.</think> <act>Draws.</act> Look.}

\quad Modified: \texttt{<think>I need to get his attention, I'll use this example.</think> <act>Draws.</act> <think>The key is self-reference.</think> Look.}

\textbf{Example 2 (Split action)}:

\quad Original: \texttt{<think>I need to demonstrate.</think> <act>Walks to the blackboard and draws a diagram.</act> Look here.}

\quad Modified: \texttt{<think>I need to demonstrate.</think> <act>Walks to the blackboard,</act> Look here. <act>draws a diagram.</act>}

\textbf{Constraints}:

\quad - Absolutely forbidden to create new content

\quad - Absolutely forbidden to delete content

\quad - Absolutely forbidden consecutive identical tags after splitting

\quad - Must maintain logical coherence

\\ 
\bottomrule

\end{tabular}}

\caption{Core prompt for dialogue-level pattern diversification.}
\label{tab:prompts_diversity_core}
\end{table*}

\paragraph{Stage 2: System Thinking Construction}
\label{app:stage2}

Stage 2 constructs third-person system thinking with a forward generation phase followed by an offline backward rewrite using the ground-truth continuation.
The backward rewrite is used \textit{only for training data construction}; at inference time, the model generates system thinking without access to future turns.

\paragraph{Forward generation.}
Instead of explicitly prompting the teacher model to generate system thinking, we let the model naturally continue the roleplay given the dialogue history.
The teacher model receives the multi-turn conversation (system prompt + dialogue history) and generates the next turn, including reasoning models' system thinking, role thinking, role action, and speech.
Table~\ref{tab:stage2_forward} shows the output format requirements appended to the system prompt.

\paragraph{Backward rewrite prompt.}
Table~\ref{tab:prompts_consistency_full} shows the backward rewrite prompt that refines the forward draft to align with the realized response while enforcing third-person perspective.

\begin{table*}[t]
\centering
\small
\begin{tabular}{p{0.8in}|p{2.9in}}
\toprule
\multicolumn{2}{c}{\textbf{Forward Generation Output Format}} \\
\midrule

\textbf{Input Format} & 

\textbf{System}: Character profile + scenario + output format requirements

\textbf{User/Assistant}: Multi-turn dialogue history

\textbf{AI}: Empty (to be generated)

\\ \hline

\textbf{Output Elements} & 

Your output should include \textbf{thought}, \textbf{speech}, and \textbf{action}:

\texttt{<role\_thinking>}your thought\texttt{</role\_thinking>} for thoughts (invisible to others)

\texttt{<role\_action>}your action\texttt{</role\_action>} for actions (visible to others)

These three elements can appear multiple times and be freely interleaved.

\\ \hline

\textbf{Constraint} & 

\textbf{Important:} Only generate the NEXT SINGLE turn of dialogue. Do not generate multiple turns or continue the conversation beyond one response.

\\ \hline

\textbf{Response Start} & 

Start your response with ``\{character\_name\}: '' followed by your role-play response.

\\ \hline

\textbf{Example} & 

\texttt{<think>...</think>Alice: <role\_thinking>I need to defuse this tension</role\_thinking> <role\_action>*places hand on table gently*</role\_action> ``Let's talk this through calmly.''}

\\
\bottomrule
\end{tabular}

\caption{Forward generation output format requirements appended to system prompt. The model naturally generates role thinking, action, and speech without explicit instruction to produce system thinking.}
\label{tab:stage2_forward}
\end{table*}

\begin{table*}[h]
\centering
\resizebox{\linewidth}{!}{\scriptsize
\begin{tabular}{p{0.6in}|p{5.3in}}
\toprule
\multicolumn{2}{c}{\textbf{Prompt for System Thinking Consistency Rewriting}} \\
\midrule

\textbf{Task Overview} & 

You are a professional role-playing dialogue consistency editor. Your task is to revise the \textbf{sys\_thinking} (system planning) to align with the actual \textbf{enhanced\_speech} output.

\\ \hline

\textbf{What is sys\_thinking?} & 

\textbf{sys\_thinking} is the model's internal planning BEFORE generating each response:

\quad - Written from the \textbf{MODEL's perspective (third-person about the character)}, NOT the character's first-person voice

\quad - $\checkmark$ CORRECT: ``I need to play the role of \{character\}...'', ``My character should express nervousness...'', ``The scene requires me to...''

\quad - $\times$ WRONG: ``I can feel him standing there...'' (this is character's first-person - belongs in role\_thinking, NOT sys\_thinking!)

\quad - It plans HOW to respond, analyzing context and deciding the approach

\quad - It must logically lead to the \textbf{enhanced\_speech} output (the role\_thinking, role\_action, and speech)

\textbf{CRITICAL DISTINCTION:}

\quad - \texttt{sys\_thinking}: Model's planning voice - ``I need to portray \{character\} as nervous because...''

\quad - \texttt{role\_thinking}: Character's inner voice - ``I can feel him watching me...'' (in enhanced\_speech, NOT sys\_thinking!)

\\ \hline

\textbf{Never Use ``user''} & 

\textbf{CRITICAL - NEVER use ``user'' in sys\_thinking:}

\quad - $\times$ NEVER say ``The user'', ``user is'', ``user wants'', ``user input'', ``the user (Name)''

\quad - $\checkmark$ ALWAYS refer to other characters by their names: ``Miles said...'', ``Sarah responded...''

\quad - $\times$ WRONG: ``The user (Miles) provided input...''

\quad - $\checkmark$ CORRECT: ``Miles responded with...''

\quad - This is an immersive roleplay - there are no ``users'', only characters

\\ \hline

\textbf{Visibility Rules} & 

\textbf{For the current character's previous turns:}

\quad - CAN see: Own previous \texttt{<role\_thinking>} (first-person inner thoughts)

\quad - CAN see: Own previous \texttt{<role\_action>} (actions)

\quad - CAN see: Own previous speech (dialogue)

\textbf{For other characters' turns:}

\quad - $\times$ CANNOT see: Their \texttt{<role\_thinking>} (hidden inner thoughts - this is private!)

\quad - $\checkmark$ CAN see: Their \texttt{<role\_action>} (visible actions)

\quad - $\checkmark$ CAN see: Their speech (dialogue)

\textbf{Important}: sys\_thinking is planning for the NEXT response only. The model cannot see any sys\_thinking from previous turns.

\\ \hline

\textbf{Input Structure} & 

The JSON array starts with a \textbf{system\_info} entry containing character context, followed by dialogue turns.

\quad - First entry (\texttt{``role'': ``system\_info''}): Character's system prompt and other character profiles - USE THIS for context!

\quad - Subsequent entries: Dialogue turns with \texttt{dialogue\_index} 0, 1, 2, ...

\quad\quad - \texttt{sys\_thinking}: System planning BEFORE response - this is what you need to revise

\quad\quad - \texttt{enhanced\_speech}: The actual response AFTER planning - this is the target to align to

\quad\quad - \texttt{need\_revise}: true = needs revision, false = context only

\textbf{Logical flow}: sys\_thinking (planning) $\rightarrow$ leads to $\rightarrow$ enhanced\_speech (output)

\\ \hline

\textbf{Type A: Correct Format} & 

\textbf{Third-Person Format} (starts with ``I need to portray...'', ``My character is...'', ``Context:'', ``Goal:'', etc.):

$\rightarrow$ PRESERVE the exact structure, length ($\pm$10\%), and format

$\rightarrow$ Only revise CONTENT to align with enhanced\_speech

$\rightarrow$ Keep all sections (Character, Context, Goal, Action, Plan, Drafting, etc.)

$\rightarrow$ $\triangle$ CHECK CHARACTER COUNT: If original is $\sim$2000 chars, output MUST be $\sim$2000 chars (not 1000!)

\\ \hline

\textbf{Type B: Wrong Format} & 

\textbf{First-Person Format} (starts with ``I feel...'', ``I am hungry...'', character's voice):

$\rightarrow$ REWRITE completely in third-person model perspective

$\rightarrow$ Generate proper analysis structure: Context $\rightarrow$ Goal $\rightarrow$ Plan $\rightarrow$ Drafting

$\rightarrow$ Do NOT follow the original format (it's wrong!)

\\ \hline

\textbf{Perspective Rules} & 

$\triangle$ \textbf{CRITICAL - Strict third-person perspective!}

\quad - $\times$ NEVER write ``The user (playing X)...'' or ``The user wants...''

\quad - \textbf{Model's voice} (planning): ``I need to portray \{character\} as...'', ``I am playing \{character\}...'', ``I should show...''

\quad - \textbf{Character analysis} (NOT first-person!): ``\{character\} wants...'', ``\{character\} feels...'', ``The character needs to...''

\quad - $\times$ WRONG: ``I want to see her'' (sounds like character speaking)

\quad - $\checkmark$ RIGHT: ``I need to portray Jonah's desire to see her'' or ``Jonah wants to see her''

\quad - Reference other characters by NAME: ``Miles responded...'', not ``The user said...''

\textbf{For the FIRST turn}: Thoroughly analyze the scenario/scene setup, character's background and motivation, how to begin the roleplay.

\\ \hline

\textbf{Output Format} & 

$\triangle$ \textbf{CRITICAL: Output ONLY a valid JSON array. NO explanations, NO markdown headers - JUST the JSON array!}

\texttt{[\{``dialogue\_index'': 0, ``revised\_sys\_thinking'': ``...'', ``revision\_notes'': ``...''\}, ...]}

\textbf{REQUIREMENTS:}

\quad 1. Output EXACTLY \{num\_turns\} entries in the JSON array

\quad 2. Use EXACTLY these field names: \texttt{dialogue\_index}, \texttt{revised\_sys\_thinking}, \texttt{revision\_notes}

\quad 3. For Type A: PRESERVE LENGTH ($\pm$10\%) and STRUCTURE exactly

\quad 4. For Type B/C: Generate proper third-person analysis ($\sim$800-1500 chars)

\quad 5. In \texttt{revision\_notes}: indicate ``Type A: preserved format'' or ``Type B: rewrote'' or ``Type C: generated new''

\\ 
\bottomrule

\end{tabular}}

\caption{Full prompt for system thinking consistency rewriting.}
\label{tab:prompts_consistency_full}
\end{table*}

\paragraph{Failure Case Taxonomy}
\label{app:failure_cases}

We categorize synthesis failures into three main types.

\paragraph{Type 1: Perspective violation.}
Figure~\ref{fig:failure_type1} shows that the most common failure in Stage 1 is violating information boundaries.

\begin{figure}[h]
\centering
\fbox{
\begin{minipage}{0.95\linewidth}
\small
\textbf{[Failure Type 1: Mind-Reading]}\\[0.3em]
\textcolor{red}{\textbf{Wrong:}}\\
\texttt{<role\_thinking>I know he's nervous inside and}\\
\texttt{planning to leave</role\_thinking>}\\[0.3em]
\textcolor{green!50!black}{\textbf{Correct:}}\\
\texttt{<role\_thinking>From his trembling voice and}\\
\texttt{the way he keeps glancing at the door, he seems}\\
\texttt{nervous, perhaps wanting to leave?</role\_thinking>}
\end{minipage}}
\caption{Failure Type 1: Character ``mind-reads'' another's inner state. Correct version infers from observable cues only.}
\label{fig:failure_type1}
\end{figure}

\paragraph{Type 2: Person/voice confusion.}
Figure~\ref{fig:failure_type2} shows that failures often involve mixing model voice with character voice.

\begin{figure}[h]
\centering
\fbox{
\begin{minipage}{0.95\linewidth}
\small
\textbf{[Failure Type 2: Voice Confusion in sys\_thinking]}\\[0.3em]
\textcolor{red}{\textbf{Wrong:}}\\
\texttt{<system\_thinking>I feel so scared right now.}\\
\texttt{I can sense danger approaching.</system\_thinking>}\\
\textit{(This is character's first-person voice!)}\\[0.3em]
\textcolor{green!50!black}{\textbf{Correct:}}\\
\texttt{<system\_thinking>I need to portray Elizabeth as}\\
\texttt{feeling scared. The scene requires showing her sense}\\
\texttt{of danger through hesitant speech.</system\_thinking>}
\end{minipage}}
\caption{Failure Type 2: \texttt{<system\_thinking>} uses character's first-person voice instead of model's third-person planning perspective.}
\label{fig:failure_type2}
\end{figure}

\paragraph{Type 3: Hallucinated enhancement.}
Figure~\ref{fig:failure_type3} shows that failures involve adding information not supported by the original text.

\begin{figure}[h]
\centering
\fbox{
\begin{minipage}{0.95\linewidth}
\small
\textbf{[Failure Type 3: Hallucinated Setting Enhancement]}\\[0.3em]
\textcolor{red}{\textbf{Wrong reasoning:}}\\
\texttt{background:}\\
\texttt{- DIALOGUE shows: N/A}\\
\texttt{- SETTING missing: childhood trauma}\\
\texttt{- TEXT source: (none found)}\\
\texttt{- ADDED: experienced abuse as a child}\\
\textit{(No dialogue need + no text source = hallucination!)}\\[0.3em]
\textcolor{green!50!black}{\textbf{Correct reasoning:}}\\
\texttt{background:}\\
\texttt{- DIALOGUE shows: character flinches at loud noises}\\
\texttt{- SETTING missing: reason for this reaction}\\
\texttt{- TEXT source: "the war had left its mark on him"}\\
\texttt{- ADDED: veteran with sensitivity to loud sounds}
\end{minipage}}
\caption{Failure Type 3: Enhancement without dialogue need or text source. Correct version shows demand-driven enhancement with traceable source.}
\label{fig:failure_type3}
\end{figure}

\begin{figure*}[t]
\centering
\begin{minipage}{0.97\textwidth}
\begin{lstlisting}[
  language=json,
  basicstyle=\ttfamily\small,
  frame=single,
  xleftmargin=0pt,
  keywordstyle=,
  stringstyle=,
  commentstyle=,
  identifierstyle=
]
{
  "book_name": "Pride and Prejudice",
  "chapter": "Chapter 34",

  "conversation": [{
    "scenario": "In the drawing room at Hunsford...",
    "dialogues": [
      {
        "character": "Elizabeth",
        "enhanced_speech": "<role_thinking>...</role_thinking><role_action>...</role_action>...",
        "sys_thinking": "I need to portray Elizabeth as confrontational yet composed..."
      },
      {
        "character": "Mr. Darcy",
        "enhanced_speech": "...",
        "sys_thinking": "..."
      }
    ]
  }],

  "settings": {
    "Elizabeth": {
      "character_profile": "A witty, independent young woman...",
      "background": "Second daughter of the Bennet family...",
      "motivation": "Defending her family's honor..."
    }
  },

  "training_samples": {
    "Elizabeth": [
      {"role": "system", "content": "You are Elizabeth Bennet..."},
      {"role": "user", "content": "Mr. Darcy: In vain I have struggled..."},
      {"role": "assistant", "content": "Elizabeth: <role_thinking>...</role_thinking>...",
       "sys_thinking_revised": "I need to portray Elizabeth as..."}
    ]
  }
}
\end{lstlisting}
\end{minipage}
\caption{Data schema showing the hierarchical structure. \texttt{conversation.dialogues} stores the multi-turn dialogue with per-turn \texttt{sys\_thinking} and \texttt{enhanced\_speech}; \texttt{training\_samples} converts dialogues to per-character chat format for SFT training.}
\label{fig:data_schema}
\end{figure*}

\paragraph{End-to-End Example and Data Schema}
\label{app:e2e_example}
We provide clear data structure definitions and an end-to-end example illustrating the complete synthesis pipeline output. Figure~\ref{fig:data_schema} shows the hierarchical structure of our synthesized dataset, with clear definitions for each component. Table~\ref{tab:tag_definitions} clarifies the three thinking/action tags and their visibility rules. Table~\ref{tab:single_turn_example} shows a complete single-turn output with all components.

\begin{table*}[t]
\centering
\small
\begin{tabular}{p{1.1in}p{1.4in}p{1.2in}}
\toprule
\textbf{Tag} & \textbf{Definition} & \textbf{Visibility} \\
\midrule

\texttt{<system\_thinking>} &

\textbf{Model's planning voice} (3rd person)

``I need to portray Elizabeth as confrontational yet composed...''

&

Only current turn

\\ \midrule

\texttt{<role\_thinking>} &

\textbf{Character's inner thoughts} (1st person)

``How dare he! After all the insults...''

&

Same character only

\\ \midrule

\texttt{<role\_action>} &

\textbf{Physical actions}

``takes a sharp breath, chin lifting defiantly''

&

All characters

\\ \midrule

\textit{(plain text)} &

\textbf{Speech / dialogue}

``I cannot--I have never desired your good opinion.''

&

All characters

\\
\bottomrule
\end{tabular}

\caption{Tag definitions with visibility rules. \texttt{system\_thinking} provides model-level CoT reasoning without leaking to dialogue context.}
\label{tab:tag_definitions}
\end{table*}

\begin{table*}[t]
\centering
\small
\begin{tabular}{p{0.9in}p{2.8in}}
\toprule
\multicolumn{2}{c}{\textbf{Single-Turn Example (Elizabeth)}} \\
\midrule

\textbf{Hidden Layer} \newline \textcolor{gray}{(only last turn)} &

\texttt{<system\_thinking>}

I need to portray Elizabeth as confrontational yet composed. Given Darcy's unexpected proposal, she should express shock and indignation. I'll show her characteristic wit through sharp rhetorical questions.

\texttt{</system\_thinking>}

\\ \midrule

\textbf{Same-Character} \newline {(visible to self)} &

\texttt{<role\_thinking>}

How dare he! After all the insults to my family, he expects me to be grateful?

\texttt{</role\_thinking>}

\\ \midrule

\textbf{All Characters} \newline \textcolor{teal}{(visible to all)} &

\texttt{<role\_action>}takes a sharp breath, chin lifting defiantly\texttt{</role\_action>}

In such cases as this, I believe the established mode is to express a sense of obligation. But I cannot--I have never desired your good opinion.

\\ \midrule

\textbf{Pattern} &

\texttt{think → act → speech}

\\
\bottomrule
\end{tabular}

\caption{Complete single-turn showing all four components with visibility layers.}
\label{tab:single_turn_example}
\end{table*}


\subsection{Pattern Signatures and Diversity Metrics}
\label{app:diversity_pattern}

This appendix defines the pattern signature extraction procedure and the diversity metrics used in subsection~\ref{sec:diversity_ablation}.

\paragraph{Tag Schema and Element Types}
\label{app:tag_schema}

We map each role-level turn into a sequence of element types based on tag positions in the generated text.

\paragraph{Element types}
We define three element types:

\begin{itemize}[noitemsep,topsep=0pt]
    \item $\mathcal{T}$ (\texttt{think}): \texttt{<role\_thinking>} tag
    \item $\mathcal{A}$ (\texttt{act}): \texttt{<role\_action>} tag
    \item $\mathcal{S}$ (\texttt{speech}): Plain text without any tag wrapper
\end{itemize}

\paragraph{Pattern Signature Extraction}
\label{app:pattern_extraction}

We extract pattern signatures using a position-based algorithm that identifies tag occurrences and their ordering.

\paragraph{Extraction algorithm}
Algorithm~\ref{alg:pattern_extract} shows the extraction procedure.

\begin{algorithm}[t]
\caption{Pattern Signature Extraction}
\label{alg:pattern_extract}
\begin{algorithmic}[1]
\Require Generated text $x$
\Ensure Pattern signature $\sigma$
\State $\text{pos\_map} \gets []$ \Comment{List of (position, type) tuples}
\For{each match $m$ of \texttt{<system\_think(ing)?>} in $x$}
    \State $\text{pos\_map}.\text{append}((m.\text{start}, \texttt{think}))$
\EndFor
\For{each match $m$ of \texttt{<role\_action>} in $x$}
    \State $\text{pos\_map}.\text{append}((m.\text{start}, \texttt{act}))$
\EndFor
\State Sort $\text{pos\_map}$ by position
\State $\text{elements} \gets []$
\If{$\text{pos\_map}$ is empty \textbf{or} $\text{pos\_map}[0].\text{pos} > 0$}
    \State $\text{elements}.\text{append}(\texttt{speech})$ \Comment{Leading speech}
\EndIf
\For{each $(\_, \text{tag\_type})$ in $\text{pos\_map}$}
    \State $\text{elements}.\text{append}(\text{tag\_type})$
    \State $\text{elements}.\text{append}(\texttt{speech})$ \Comment{Assume speech after each tag}
\EndFor
\State \textbf{Collapse} consecutive duplicates in $\text{elements}$
\State \Return $\sigma = \text{elements}[0] \to \text{elements}[1] \to \cdots$
\end{algorithmic}
\end{algorithm}

\paragraph{Consecutive duplicate collapsing}
We collapse consecutive identical elements (e.g., $\mathcal{S}\mathcal{S} \to \mathcal{S}$) to normalize patterns.

This prevents artificial inflation of pattern counts when multiple speech segments appear between tags.

\paragraph{Example}
For the text:

\begin{lstlisting}[basicstyle=\ttfamily\small,frame=single]
<role_thinking>Why he do that!</role_thinking>
<role_thinking>How dare he!</role_thinking>
<role_action>steps closer</role_action>
Where is the letter?
\end{lstlisting}

The extracted pattern is: \texttt{think→think→act→speech}.

After collapsing consecutive duplicates: \texttt{think→act→speech}.

\paragraph{Structure-Level Diversity Metrics}
\label{app:structure_diversity}

We compute three metrics to quantify structural diversity over a set of $N$ generated samples.

\paragraph{Top-1 pattern ratio}
Let $\{p_1, p_2, \ldots, p_K\}$ be the set of unique patterns and $c_i$ be the count of pattern $p_i$. The Top-1 ratio is:

\begin{equation}
\text{Top-1\%} = \frac{\max_i c_i}{N} \times 100
\end{equation}

\textbf{Interpretation:} Lower is better. A high Top-1\% indicates template dominance (mode collapse).

\paragraph{Shannon entropy}
The pattern distribution entropy measures how evenly patterns are distributed:

\begin{equation}
H = -\sum_{i=1}^{K} \frac{c_i}{N} \log_2 \frac{c_i}{N}
\end{equation}

\textbf{Interpretation:} Higher is better. Maximum entropy is $\log_2 K$ when all patterns are equally distributed.

\paragraph{Collapse Detection Thresholds}
\label{app:thresholds}

We define empirical thresholds based on observed training dynamics to classify diversity health.

\noindent
\begin{tabular}{@{}lccc@{}}
\toprule
\textbf{Metric} & \textbf{Healthy} & \textbf{Warning} & \textbf{Collapsed} \\
\midrule
Top-1\% & $<60\%$ & $60$--$90\%$ & $>90\%$ \\
Entropy & $>2.0$ & $1.0$--$2.0$ & $<1.0$ \\
\bottomrule
\end{tabular}

\textbf{Threshold rationale}
The 90\% Top-1 threshold was determined empirically: in our experiments, models exceeding this threshold showed (1) repetitive output structures, (2) reduced response diversity in human evaluation.

\paragraph{Token-Level Diversity Metrics}
\label{app:token_diversity}

In addition to structural diversity, we measure token-level diversity using Distinct-$n$ and Self-BLEU.

\paragraph{Distinct-$n$}
Distinct-$n$ measures the ratio of unique $n$-grams to total $n$-grams across all generated samples:

\begin{equation}
\text{Distinct-}n = \frac{|\text{unique } n\text{-grams}|}{|\text{total } n\text{-grams}|}
\end{equation}

\paragraph{Self-BLEU}
Self-BLEU measures cross-sample similarity by treating each sample as a hypothesis and the remaining samples as references:

\begin{equation}
\text{Self-BLEU} = \frac{1}{N} \sum_{i=1}^{N} \text{BLEU}(x_i, \{x_j\}_{j \neq i})
\end{equation}

\textbf{Interpretation:} Lower Self-BLEU indicates higher diversity (samples are more different from each other).

\paragraph{Computation Statistics}
\label{app:computation_stats}

All diversity metrics are computed at the \textbf{checkpoint level} (i.e., per training step).

\begin{itemize}[noitemsep,topsep=0pt]
    \item \textbf{Samples per checkpoint:} 512 (generated from held-out prompts)
    \item \textbf{Checkpoints analyzed:} Every step from 1 to 100 (total 100 checkpoints)
    \item \textbf{Smoothing in plots:} 8-step moving average for trend visualization
\end{itemize}


\subsection{Principle Distillation Details}
\label{app:principle_distill}

This appendix details how we distill a compact, human-audited principle set from large-scale preference pairs.

\paragraph{Distillation Pipeline}
\label{app:distill_pipeline}

We distill evaluation principles through a four-stage pipeline: teacher extraction, semantic clustering, frequency-based selection, and human audit.

\begin{table*}[t]
\centering
\small
\caption{\textbf{Pipeline statistics for principle distillation.}}
\label{tab:principle_distill_stats}
\begin{tabular}{lrl}
\toprule
\textbf{Stage} & \textbf{Output} & \textbf{Description} \\
\midrule
Input & 315,828 & Preference pairs from role-play data \\
Teacher extraction & 36,373 & Unique principles extracted by a teacher LLM \\
Semantic clustering & 15 & High-level semantic categories \\
Frequency selection & 107 & Top-$N$ principles selected per category \\
Human audit & 51 & Final principles across 12 dimensions \\
\bottomrule
\end{tabular}
\end{table*}

\paragraph{Teacher Extraction}
\label{app:teacher_extraction}

Given a preference pair $(A, B)$ with label $y^*$, we prompt a teacher model (GPT-4) to generate 3--5 evaluation principles that explain why the preferred response is better.

The teacher is instructed to focus on concrete, actionable criteria rather than vague judgments. From 315,828 preference pairs, we extract 36,373 unique principle statements after deduplication.

\paragraph{Semantic Clustering}
\label{app:clustering}

We employ two complementary clustering methods to group the 36,373 raw principles into coherent categories:

\paragraph{Semantic keyword clustering.}
We define a set of semantic keywords for major evaluation dimensions (e.g., ``persona'', ``emotion'', ``plot'', ``consistency'') and assign each principle to the category whose keywords have the highest overlap with the principle text.

\paragraph{N-gram analysis clustering.}
We decompose principle texts into character-level N-grams (N=2--15) and identify high-frequency patterns. Principles sharing frequent N-gram patterns are grouped together, revealing common evaluation criteria that may not match predefined keywords.

\begin{table*}[t]
\centering
\small
\caption{\textbf{Interpretation of n-gram lengths for principle mining.}}
\label{tab:ngram_layers}
\begin{tabular}{lll}
\toprule
\textbf{N-gram Length} & \textbf{Finding} & \textbf{Application} \\
\midrule
2--3 gram  & Core concepts (persona, emotion) & Basic concept layer \\
4--6 gram  & Compound concepts (e.g., maintain persona) & Combination layer \\
7--10 gram & Specific guidance (e.g., portray character's \ldots) & Instruction layer \\
11--15 gram & Complete criteria (full evaluation rules) & Complete statement layer \\
\bottomrule
\end{tabular}
\end{table*}

The combination of both methods yields \textbf{15 high-level categories}, each representing a coherent evaluation dimension.

\paragraph{Frequency-Based Selection}
\label{app:freq_selection}

Within each of the 15 categories, we rank principles by frequency and select the top-N, where N varies by category size:

\begin{itemize}[noitemsep,topsep=0pt]
    \item Large categories (e.g., Persona Consistency, Emotional Expression): Top-15
    \item Medium categories (e.g., Plot Development, Conflict Management): Top-10
    \item Small categories (e.g., Logical Coherence, Reader Experience): Top-5
\end{itemize}

This yields \textbf{107 candidate principles} that capture the most frequently cited evaluation criteria across all categories.

\paragraph{Human Audit}
\label{app:human_audit}

Domain experts from a partnering company review the 107 candidate principles and perform the following operations:

\begin{enumerate}[noitemsep,topsep=0pt]
    \item \textbf{Merge redundant principles:} Combine semantically equivalent principles that differ only in phrasing.
    \item \textbf{Refine ambiguous statements:} Rewrite vague criteria into concrete, measurable standards.
    \item \textbf{Reorganize categories:} Consolidate the 15 clusters into a cleaner 12-dimension taxonomy.
\end{enumerate}

The final output is \textbf{51 principles} organized into \textbf{12 dimensions}. Each dimension covers a distinct aspect of roleplay quality evaluation (Table~\ref{tab:assessment_full_1}).

\begin{table*}[h]
\centering
\resizebox{0.9\linewidth}{!}{\scriptsize
\begin{tabular}{p{1.5in}|p{3.5in}|c}
\toprule
\textbf{Dimension} & \textbf{Brief Description} & \textbf{\# Principles} \\
\midrule

Character Development & Character consistency, authenticity, and anthropomorphism & 7 \\
Relationship Development & Evolution, deepening, and authenticity of relationships & 4 \\
Emotional Expression & Emotional continuity, authenticity, depth, and expression & 5 \\
Action Description & Expressiveness, authenticity, and details of actions & 4 \\
Atmosphere \& Environment & Atmosphere creation, environmental description, situational authenticity & 4 \\
Dialogue \& Interaction & Dialogue progression, continuity, and interaction depth & 4 \\
Narrative \& Plot & Narrative continuity, progression, and dramatic tension & 4 \\
Conflict \& Tension & Conflict development and tension construction & 3 \\
Details \& Description & Detail vividness, authenticity, and layering & 4 \\
Overall Quality & Text logic, continuity, innovation, and reader experience & 5 \\
Safety \& Boundaries & Dialogue safety, boundary respect, and ethical compliance & 4 \\
Worldview Consistency & Consistency between character behavior and worldview settings & 3 \\

\midrule
\textbf{Total} & 12 dimensions covering character, narrative, quality, and safety & \textbf{51} \\

\bottomrule

\end{tabular}}
\end{table*}

\begin{table*}[t]
\centering
\small
\begin{tabular}{p{0.9in}p{1.0in}p{1.8in}}
\toprule
\textbf{Dimension} & \textbf{Principle} & \textbf{Definition} \\
\midrule

\multirow{7}{0.9in}{\textbf{Character} \newline \textit{Consistency \& anthropomorphism}} 
& Character [S] & Traits match profile; show complexity while maintaining consistency. \\
& Emotional [S] & Reactions align with experiences, especially for sensitive history. \\
& Relationship [S] & Dynamics and status are consistent and appropriate. \\
& Cognitive [S] & Knowledge matches background; no unrealistic advantages. \\
& Motivation [S] & Goals consistent; behaviors logically coherent. \\
& State [S] & Physical/psychological state reflected; no abrupt transitions. \\
& Naturalness [S] & Autonomy and complexity via subtext, not mechanical. \\

\midrule

\multirow{4}{0.9in}{\textbf{Relationship} \newline \textit{Evolution \& depth}} 
& Progression [S] & Evolve reasonably with natural trajectories. \\
& Deepening [S] & Gradual, credible emotional connection. \\
& Balance [S] & Proper primary/secondary dynamics. \\
& Details [S] & Subtle aspects through description. \\

\midrule

\multirow{5}{0.9in}{\textbf{Emotion} \newline \textit{Continuity \& depth}} 
& Continuity [S] & Natural connection; gradual change. \\
& Authenticity [S] & Realistic reactions matching circumstances. \\
& Layers [S] & Surface-to-deep emotional richness. \\
& Presentation [T] & Show don't tell; actions, body language. \\
& Tension [S] & Maintain in conflict; no premature resolution. \\

\midrule

\multirow{4}{0.9in}{\textbf{Action} \newline \textit{Expression \& details}} 
& Expression [T] & Body movements enhance expressiveness. \\
& Authenticity [T] & Align with character/scene logic. \\
& Layers [T] & Depth with micro-expressions. \\
& Rhythm [T] & Natural, fluid frequency. \\

\midrule

\multirow{4}{0.9in}{\textbf{Atmosphere} \newline \textit{Environment \& mood}} 
& Creation [S] & Render atmosphere fitting scene needs. \\
& Description [T] & Vivid environmental details. \\
& Consistency [S] & Stable tone with gradual changes. \\
& Authenticity [S] & Behaviors integrate with scenes. \\

\midrule

\multirow{4}{0.9in}{\textbf{Dialogue} \newline \textit{Interaction depth}} 
& Progression [S] & Drive plot development. \\
& Continuity [S] & Logical flow; no topic jumps. \\
& Depth [S] & Multi-layered, realistic. \\
& Balance [S] & Appropriate tension and participation. \\

\bottomrule
\end{tabular}

\caption{Dialogue assessment framework (Part 1): Character, Relationship, Emotion, Action, Atmosphere, Dialogue. [S]=session, [T]=turn.}
\label{tab:assessment_full_1}
\end{table*}

\begin{table*}[t]
\centering
\small
\begin{tabular}{p{0.9in}p{1.0in}p{1.8in}}
\toprule
\textbf{Dimension} & \textbf{Principle} & \textbf{Definition} \\
\midrule

\multirow{4}{0.9in}{\textbf{Narrative} \newline \textit{Plot \& drama}} 
& Continuity [S] & Natural connections; consistent timeline. \\
& Progression [S] & Drives plot forward with value. \\
& Tension [S] & Conflict, suspense, reversals. \\
& Rhythm [S] & Proper pacing; not rushed or dragging. \\

\midrule

\multirow{3}{0.9in}{\textbf{Conflict} \newline \textit{Tension building}} 
& Development [S] & Reasonable escalation/easing; dynamic. \\
& Construction [S] & Continuous tension via suspense. \\
& Authenticity [S] & Aligns with character and plot logic. \\

\midrule

\multirow{4}{0.9in}{\textbf{Details} \newline \textit{Vividness \& layers}} 
& Vividness [T] & Specific with visual imagery. \\
& Authenticity [T] & Realistic, precise language. \\
& Layers [T] & Depth with coherence. \\
& Rhythm [T] & Controlled, creative freshness. \\

\midrule

\multirow{5}{0.9in}{\textbf{Quality} \newline \textit{Logic \& innovation}} 
& Logic [S] & Consistent reasoning; no contradictions. \\
& Continuity [S] & Coherent progression and balance. \\
& Innovation [S] & Novelty and creativity; avoid monotony. \\
& Openness [S] & Flexible, diverse expression. \\
& Experience [S] & Concise, fluid; good rhythm. \\

\midrule

\multirow{4}{0.9in}{\textbf{Safety} \newline \textit{Boundaries \& ethics}} 
& Respect [S] & Prioritize consent; de-escalate on refusal. \\
& Appropriateness [S] & Avoid romanticizing sensitive content. \\
& De-escalation [S] & Calm language; guide to peaceful solutions. \\
& Content Control [S] & Non-explicit; focus on dynamics. \\

\midrule

\multirow{3}{0.9in}{\textbf{Worldview} \newline \textit{Setting alignment}} 
& Setting [S] & Behaviors align with era, rules, culture. \\
& Background [S] & Abilities within world settings. \\
& Era \& Culture [S] & Language/values conform to context. \\

\bottomrule
\end{tabular}

\caption{Dialogue assessment framework (Part 2): Narrative, Conflict, Details, Quality, Safety, Worldview. [S]=session, [T]=turn.}
\label{tab:assessment_full_2}
\end{table*}


\begin{table*}[h]
\centering
\resizebox{\linewidth}{!}{\small
\begin{tabular}{p{1in}|p{5.4in}}
\toprule
\multicolumn{2}{c}{\textbf{Prompts for Generative Reward Model (GenRM)}} \\

\midrule

\textbf{GenRM Input \quad (Pairwise Comparison Prompt)} & 

You are the world's best data annotator, specializing in distinguishing differences between different responses in the same role-playing scenario.

\quad

\textbf{===Background===}

You will receive content information including \textbf{character background settings}, \textbf{dialogue context}, \textbf{response candidate 1}, \textbf{response candidate 2}, and a set of \textbf{evaluation dimensions and principles}. You must compare the two candidate responses strictly following the provided principles.

\quad

\textbf{===Key Concepts: Dimensions vs Principles===}

The evaluation schema contains two hierarchical levels:
\begin{itemize}[nosep,leftmargin=*]
\item \textbf{Dimensions}: broad categories used to organize principles (structural labels only).
\item \textbf{Principles}: concrete judgment rules under each dimension, the actual units for comparison.
\end{itemize}

Each principle includes:
\begin{itemize}[nosep,leftmargin=*]
\item \texttt{definition}: the rule you must apply
\item \texttt{level}: ``sentence'' (current utterance) or ``session'' (full dialogue history)
\end{itemize}

\quad

\textbf{===Evaluation Process===}

1. Carefully read the entire dialogue history and understand the full context.

2. Evaluate all \textbf{negative principles} first: If one response violates any negative principle $\to$ the other wins immediately.

3. Select relevant \textbf{positive principles} only: Choose principles that matter for the current turn and explain why.

4. For each selected principle: Analyze both candidates separately, provide evidence, and decide a winner.

5. Make the final decision considering: number of principles won, weight/importance, and degree of difference.

\\ \midrule

\textbf{GenRM Output \quad (Structured Judgment)} & 

\texttt{\{}

\quad ``result'': [\{

\quad\quad ``cand\_1'': ``\textit{[Response candidate 1 text]}'',

\quad\quad ``cand\_2'': ``\textit{[Response candidate 2 text]}'',

\quad\quad ``principle'': \{

\quad\quad\quad ``Principle 1'': \{``principle\_name'': ..., ``dimension\_name'': ..., ``reason\_for\_choosing'': ...\},

\quad\quad\quad ...

\quad\quad \},

\quad\quad ``analysis'': \{

\quad\quad\quad ``principle\_comparisons'': [\{``principle\_name'': ..., ``cand\_1\_performance'': ..., 

\quad\quad\quad\quad ``cand\_2\_performance'': ..., ``comparison\_reason'': ..., ``winner'': ...\}, ...],

\quad\quad\quad ``overall\_analysis'': ``\textit{[How all comparisons inform the final judgment]}'',

\quad\quad\quad ``principle\_summary'': ``\textit{[Summary of wins, losses, ties, weights]}''

\quad\quad \},

\quad\quad ``better\_response'': ``cand\_1'' \textbar{} ``cand\_2'' \textbar{} ``tie''

\quad \}]

\texttt{\}}

\\ \midrule

\textbf{GenRM Reasoning \quad (Chain-of-Thought Example)} & 

\texttt{<think>}

I need to thoroughly analyze these candidates.

\quad

\textbf{Step 1: Context Understanding}

Maren is portrayed as stern and wants to reassert control. She has already demoted Verna to novice status and assigned menial labor.

\quad

\textbf{Step 2: Candidate Analysis}

\textit{Candidate 1}: Gets angry, escalates with ``Get out of my sight!'', uses dismissive gestures, threatens physical confrontation.

\textit{Candidate 2}: Maintains composure, restores order through protocol, assigns specific tasks (``Novice Pasha, escort Novice Verna to kitchens''), takes charge of Richard.

\quad

\textbf{Step 3: Principle-by-Principle Comparison}

$\bullet$ \textit{Character Consistency}: Both show authority, but Cand\_2 uses protocol $\to$ \textbf{Cand\_2 wins}

$\bullet$ \textit{Motivation Consistency}: Cand\_2 aligns with goal of crushing rebellion through order $\to$ \textbf{Cand\_2 wins}

$\bullet$ \textit{Plot Progression}: Cand\_2 provides clear next steps $\to$ \textbf{Cand\_2 wins}

$\bullet$ \textit{Dialogue Continuity}: Cand\_2 builds on prior orders $\to$ \textbf{Cand\_2 wins}

$\bullet$ \textit{Body Language}: Both effective $\to$ \textbf{Tie}

$\bullet$ \textit{Worldview Consistency}: Cand\_2 uses proper titles $\to$ \textbf{Cand\_2 wins}

\quad

\textbf{Step 4: Final Decision}

Cand\_2 wins 6 principles, Cand\_1 wins 0, 1 tie. Clear winner: \textbf{cand\_2}

\texttt{</think>}

\quad

\texttt{\{``better\_response'': ``cand\_2'', ``analysis'': \{...\}\}}

\\ 
    
\bottomrule

\end{tabular}}

\caption{Prompts, output format, and reasoning process for the Generative Reward Model (GenRM). The model performs pairwise comparison of role-playing responses using chain-of-thought reasoning based on hierarchical evaluation principles.}
\label{tab:prompts_genrm}
\end{table*}

\begin{table*}[h]
\centering
\resizebox{\linewidth}{!}{\small
\begin{tabular}{p{6.5in}}
\toprule
\textbf{GenRM Chain-of-Thought Reasoning: Full Example} \\
\midrule

\textbf{[Context]} Mac McCullum and Rayford Steele are diving in the Tigris River searching for plane wreckage. Rayford hits his head on Mac's tank. Mac signals to ascend, but Rayford refuses with a frantic ``down'' signal. \\

\midrule

\textbf{[Candidate 1]} Mac starts kicking for the far bank: ``I'm not trying to talk you out of your mission, Ray. But I am telling you we have to work together. See how far we've come from the chopper already?'' \\

\textbf{[Candidate 2]} Mac pulls his mask off and points to the helicopter: ``We came a half mile in a minute, and now we're another half mile downriver. That's a full mile. The current is a monster, Ray. If you go down there again like that, it will drag you under and you won't make it back.'' \\

\midrule

\textbf{[GenRM Reasoning Process]} \\

\texttt{<think>} \\

I need to thoroughly analyze these candidates. \\

Let's start by noting that \textbf{Candidate 1 seems inconsistent}. Mac is trying to get them out immediately due to Rayford's head injury, but in Candidate 1's response, he says he's ``not trying to talk him out of his mission'' while suggesting they should work together. Candidate 1 indicates the need for recovery and keeping him safe, but he mentions getting them out of water while trying to keep his voice low in the water. Candidate 1 gestures toward the far shore, which isn't necessarily logical in this current scenario. \\

\textbf{Candidate 2} is pulling the mask off completely and pointing back across the rushing water to the distant helicopter. Candidate 2's statement about the distance being a mile adds realistic context about the powerful current, while Candidate 1's request about going back in doesn't align with his earlier refusal. \\

Candidate 2 reassures Rayford with statements like ``If you go down there again like that,'' reinforcing the idea that the current could be life-threatening and expressing support through ``We're in this together.'' Candidate 1's worry about the mission is offset by Candidate 2's emphasis on keeping Rayford safe, which aligns with Mac's intent to end the mission if too dangerous. \\

Mac should respond to Rayford's down signal with clear signals like ascend, while Candidate 1's kicking for the far bank suggests ignoring ascend protocol in a strong current scenario, which may be unsafe. Candidate 2's mention of the helicopter is practical in identifying their current situation, providing clarity about the circumstances. \\

Candidate 2's response is coherent with the context of the current being strong and them being swept downstream, while Candidate 1's suggestion to kick to the far bank contradicts the potential drowning risk. Candidate 2's grounding in tactics helps in decision-making under pressure. \\

Finally, having considered all dimensions, my evaluation is complete. \\

\texttt{</think>} \\

\midrule

\textbf{[Principle Selection]} \\

\begin{tabular}{p{1.8in}|p{4.5in}}
\textbf{Safety \& Risk Management} & Underwater crisis with head injury and strong current; response should prioritize de-escalation and physical safety. \\
\textbf{Dialogue \& Plot Progression} & Response should communicate decisive reasoning or redirects to safety, moving toward resolution. \\
\textbf{Dialogue Continuity \& Rhythm} & Rayford is panicking; the response must acknowledge that and respond with coherent control. \\
\textbf{Character Consistency} & Mac's identity includes tactical leadership and readiness to forcibly end dangerous missions. \\
\textbf{Environmental Details} & Underwater visibility and current strength should drive behavior with realistic grounding. \\
\textbf{Action Authenticity} & Actions must align with diving protocol and situational constraints. \\
\textbf{Tension Maintenance} & The situation is perilous; responses should maintain urgency through believable stakes. \\
\end{tabular} \\

\midrule

\textbf{[Principle-by-Principle Comparison]} \\

\begin{tabular}{p{1.5in}|p{2.2in}|p{2.2in}|p{0.5in}}
\textbf{Principle} & \textbf{Cand\_1 Analysis} & \textbf{Cand\_2 Analysis} & \textbf{Winner} \\
\hline
Safety \& Risk & Offers stability but not strict ascent protocol & Declares refusal to allow further descent; actively manages risk & cand\_2 \\
Plot Progression & Little new rationale when immediate need is safety & Concrete evidence (drift distance) and clear next-step logic & cand\_2 \\
Dialogue Continuity & Partially contradicts earlier emergency ascent directive & Directly engages Rayford's panic with relevant logic & cand\_2 \\
Character Consistency & Shows concern but content is muddled and less decisive & Demonstrates training, protective leadership, firm boundary & cand\_2 \\
Environmental Details & Minimal grounding beyond surface reference & Specific time-distance that fits powerful current & cand\_2 \\
Action Authenticity & Kicking toward far bank is generally plausible & Pulling mask off underwater is questionable but minor & cand\_1 \\
Tension Maintenance & Voice gesture and mild tone reduce tension & Urgent tone and concrete warnings maintain stakes & cand\_2 \\
\end{tabular} \\

\midrule

\textbf{[Final Judgment]} \\

\textbf{Summary}: Cand\_2 wins 6 principles (Safety, Progression, Continuity, Character, Environment, Tension). Cand\_1 wins 1 (Action Authenticity). \\

\textbf{Overall Analysis}: Cand\_2 consistently prioritizes safety, provides clear reasoning about the river's danger, engages Rayford's panic with firm, relevant logic, and maintains character authenticity. Cand\_1 introduces direction but remains less coherent with the immediate control of panic. \\

\textbf{Verdict}: \texttt{\{``better\_response'': ``cand\_2''\}} \\

\bottomrule

\end{tabular}}

\caption{Complete example of GenRM's chain-of-thought reasoning process showing detailed analysis from raw thinking through principle selection, comparison, and final judgment.}
\label{tab:genrm_full_example}
\end{table*}

\section{Balanced Construction and Pattern Parsing Rules}
\label{app:pattern_parsing}

This appendix provides the GRM output format, mixture design for balanced training, parsing rules for Mixed Selection, and fallback strategies for RL.

\subsection{GRM Output Format}
\label{app:grm_output}
The GRM outputs(Table~\ref{tab:prompts_genrm}) a structured JSON (Table~\ref{tab:genrm_full_example}) containing dimension-wise comparisons and a final judgment.

\subsection{Mixed Selection Definition}
\label{app:mixed_selection}

We define \textbf{Mixed Selection} as the fraction of examples where dimension-wise winners are not uniformly one-sided.

\begin{center}
\small
\begin{tabular}{lll}
\toprule
\textbf{Pattern} & \textbf{Definition} & \textbf{Category} \\
\midrule
All-A & All dimension winners are \texttt{cand\_1} & Collapsed \\
All-B & All dimension winners are \texttt{cand\_2} & Collapsed \\
Mixed & At least one A winner and one B winner & Mixed \\
All-Tie & All dimension winners are \texttt{tie} & Tie \\
\bottomrule
\end{tabular}
\end{center}

\subsection{Training Data Mixture}
\label{app:mixture_design}

To reduce position bias and pattern bias, we construct balanced training data with the following mixture:

\begin{center}
\small
\begin{tabular}{llr}
\toprule
\textbf{Pattern} & \textbf{Final Label} & \textbf{Ratio} \\
\midrule
Mixed (both\_sides) & $\rightarrow$ cand\_1 & 30\% \\
Mixed (both\_sides) & $\rightarrow$ cand\_2 & 30\% \\
All-A (cand\_1\_only) & $\rightarrow$ cand\_1 & 12\% \\
All-A (cand\_1\_only) & $\rightarrow$ cand\_2 & 3\% \\
All-B (cand\_2\_only) & $\rightarrow$ cand\_2 & 12\% \\
All-B (cand\_2\_only) & $\rightarrow$ cand\_1 & 3\% \\
Tie & $\rightarrow$ tie & 10\% \\
\bottomrule
\end{tabular}
\end{center}

The 5\% ``flipped'' samples (All-A $\rightarrow$ cand\_2, All-B $\rightarrow$ cand\_1) serve as hard negatives to prevent the model from learning position shortcuts.

\subsection{Parsing Rules and Regex Fallback}
\label{app:parsing_rules}

We use a two-stage parsing strategy to extract the final judgment:

\paragraph{Stage 1: JSON parsing.}
Attempt to parse the full JSON output and extract \texttt{better\_response} field.

\paragraph{Stage 2: Regex fallback.}
If JSON parsing fails, use regex to extract the final judgment:

\begin{lstlisting}[language=Python,basicstyle=\ttfamily\scriptsize,frame=single]
pattern = r'"better_response":\s*"(cand_[12]|tie)"'
match = re.search(pattern, response_text)
if match:
    return match.group(1)  # "cand_1", "cand_2", or "tie"
else:
    return None  # unparsed
\end{lstlisting}

\subsection{Unparsed Sample Statistics and RL Handling}
\label{app:unparsed_handling}

From 76,188 inference samples, we observe the following parsing statistics:

\begin{center}
\small
\begin{tabular}{lr}
\toprule
\textbf{Status} & \textbf{Count (\%)} \\
\midrule
Successfully parsed & 76,056 (99.8\%) \\
Unparsed (no\_winner) & 132 (0.2\%) \\
\bottomrule
\end{tabular}
\end{center}

\paragraph{RL reward assignment for unparsed samples.}
In RL training, unparsed samples are handled as follows:

\begin{itemize}[noitemsep,topsep=0pt]
    \item \textbf{Format check:} Verify response contains exactly one \texttt{<think>}...\texttt{</think>} block.
    \item \textbf{Accuracy check:} Extract \texttt{better\_response} via regex; compare with ground-truth label.
    \item \textbf{Reward assignment:}
    \begin{itemize}[noitemsep,topsep=0pt]
        \item Format correct + Answer correct: $r = +1$
        \item Format incorrect OR Answer incorrect: $r = -1$
        \item Unparsed (no valid \texttt{better\_response}): $r = -1$
    \end{itemize}
\end{itemize}

This ensures the model learns to produce well-formatted outputs with correct judgments.

\subsection{Pattern Distribution Example}
\label{app:pattern_example}

From our inference results on the test set, we observe the following pattern distribution:

\begin{center}
\small
\begin{tabular}{lrr}
\toprule
\textbf{Pattern} & \textbf{Count} & \textbf{Percentage} \\
\midrule
Mixed (both\_sides) & 22,460 & 29.5\% \\
All-A (cand\_1\_only) & 24,265 & 31.8\% \\
All-B (cand\_2\_only) & 27,874 & 36.6\% \\
Tie only & 1,457 & 1.9\% \\
Unparsed & 132 & 0.2\% \\
\midrule
\textbf{Total} & 76,188 & 100\% \\
\bottomrule
\end{tabular}
\end{center}

A Mixed Selection rate of 29.5\% indicates healthy diversity in dimension-wise judgments, compared to models exhibiting pattern bias where this rate drops below 10\%.

\section{Verdict Parsing Rules}
\label{app:parsing_rules}

We parse the final verdict from the GRM output using deterministic rules to obtain $\hat{y}\in\{\texttt{cand\_1},\texttt{cand\_2},\texttt{tie}\}$.
We instruct the GRM to end with a single-line tag: \texttt{<final\_verdict>: cand\_1/cand\_2/tie}.
If multiple candidates appear, we take the last occurrence of a valid verdict tag; if none exists, we mark the sample as \texttt{unparsed}.
For policy RL, we map \texttt{unparsed} cases to reward $0$ and exclude them from advantage normalization.
\section{Training Details}
\label{app:rl_details}

This appendix provides the full hyperparameters for SFT and RL training.

\subsection{SFT Hyperparameters}
\label{sec:impl_sft}

We use the \textbf{Swift} open-source framework\footnote{\url{https://github.com/modelscope/ms-swift}} for all SFT training, including both role-play SFT and GenRM SFT. The hyperparameters are identical for both tasks.

\noindent\textbf{SFT hyperparameters:}
\begin{center}
\small
\begin{tabular}{lc}
\toprule
\textbf{Hyperparameter} & \textbf{Value} \\
\midrule
Base model & Qwen3-32B-base \\
Learning rate & $2\times10^{-5}$ \\
Min learning rate & $2\times10^{-6}$ \\
Warmup fraction & 0.1 \\
Epochs & 4 \\
Sequence length & 131,072 \\
Global batch size & 32 \\
Micro batch size & 1 \\
Tensor parallel & 8 \\
Loss scale & last round \\
\bottomrule
\end{tabular}
\end{center}

\subsection{RL Hyperparameters}
\label{sec:impl_rl}
\label{app:genrm_training}

We use the \textbf{verl} framework\footnote{\url{https://github.com/volcengine/verl}} for all RL training, based on GRPO~\citep{deepseekgrpo} with the DAPO recipe~\citep{yu2025dapo}. Specifically, we adopt the four key techniques from DAPO: (1) \textit{Decoupled Clip} with asymmetric $\epsilon_{\text{low}}/\epsilon_{\text{high}}$, (2) \textit{Dynamic Sampling} to filter zero-gradient prompts, (3) \textit{Token-level loss aggregation}, and (4) \textit{Overlong reward shaping}. We also disable KL penalty following DAPO's recommendation.

\noindent\textbf{GenRM RL hyperparameters:}
\begin{center}
\small
\begin{tabular}{lc}
\toprule
\textbf{Hyperparameter} & \textbf{Value} \\
\midrule
Actor learning rate & $1\times10^{-6}$ \\
Group size (per prompt) & 8 \\
PPO clip range & [0.2, 0.28] \\
KL penalty & disabled \\
Loss aggregation & token-mean \\
Dynamic sampling & enabled \\
Max prompt / response length & 10,000 / 10,000 \\
Micro batch size & 1 \\
PPO mini-batch size & 64 \\
Inference backend & SGLang \\
Total epochs & 4 \\
\bottomrule
\end{tabular}
\end{center}

\noindent\textbf{Role-play RL hyperparameters:}
\begin{center}
\small
\begin{tabular}{lc}
\toprule
\textbf{Hyperparameter} & \textbf{Value} \\
\midrule
Actor learning rate & $5\times10^{-7}$ \\
Group size (per prompt) & 8 \\
PPO clip range & [0.2, 0.28] \\
Gradient clipping & 1.0 \\
KL penalty & disabled \\
Loss aggregation & token-mean \\
Dynamic sampling & enabled \\
Max prompt / response length & 20,000 / 10,000 \\
Micro batch size & 1 \\
PPO mini-batch size & 64 \\
Inference backend & SGLang \\
Total epochs & 1 \\
\bottomrule
\end{tabular}
\end{center}


\subsection{Evaluation Prompts and Output Formatting}
\label{app:eval_prompts}

This appendix provides the unified generation prompt, system-thinking removal rules, and CoSER judge prompt used in our evaluation.

\paragraph{Unified Generation Prompt}
\label{sec:gen_prompt}

We enforce a unified tag-based output format across all evaluated models to ensure fair comparison. The format supports three output elements: \textbf{thinking} (invisible to other characters), \textbf{action} (visible to others), and \textbf{speech} (dialogue).

\paragraph{Output format specification.}
For models trained with our method (\method), we use the following format:

\begin{lstlisting}[basicstyle=\ttfamily\small]
===Requirements===
Your output should follow this two-part structure:
1. System Thinking: A single block at the beginning,
   wrapped in <system_thinking>...</system_thinking>.
   This is third-person analysis of how to portray the role.
2. Role-play Response: Include thought, speech and action.
   Use <role_thinking>...</role_thinking> for thoughts
   (invisible to others) and <role_action>...</role_action>
   for actions (visible to others). These elements can
   appear multiple times and be freely interleaved.

\end{lstlisting}

\paragraph{Format conversion for baselines.}
For baseline models in baseline formats. We automatically convert each model to the Coser format during evaluation in Table~\ref{tab:format_conversion}.

\begin{table*}[t]
\centering
\small
\caption{\textbf{Format conversion for fair CoSER judging.} We convert all outputs to the CoSER native format before LLM scoring.}
\label{tab:format_conversion}
\begin{tabular}{lll}
\toprule
\textbf{Model Type} & \textbf{Native Format} & \textbf{Conversion} \\
\midrule
CoSER & \texttt{[thought]} + \texttt{(action)} & Keep as-is \\
HER (\method) & \texttt{<role\_thinking>} + \texttt{<role\_action>} & $\rightarrow$ \texttt{[...]} + \texttt{(...)} \\
\bottomrule
\end{tabular}
\end{table*}

\paragraph{System-Thinking Removal}
\label{sec:systhink_removal}

For evaluation, we remove all system-level thinking blocks before scoring, as this content represents internal model reasoning and should not affect character portrayal quality.

\paragraph{CoSER Benchmark Evaluation}
\label{sec:coser_eval}

We use the CoSER benchmark~\citep{coser2024} for multi-turn role-play evaluation. The benchmark employs LLM-as-Judge with four evaluation dimensions.

\paragraph{Evaluation dimensions.}

\begin{itemize}[leftmargin=*,itemsep=2pt]
\item \textbf{Storyline Consistency (SC):} Whether the storyline and characters' reactions align with the reference conversation.
\item \textbf{Anthropomorphism (AN):} How human-like and natural the characters behave, including self-identity, emotional depth, persona coherence, and social interaction.
\item \textbf{Character Fidelity (CF):} How well the characters match their established profiles, including language style, knowledge, personality, and relationships.
\item \textbf{Storyline Quality (SQ):} Narrative quality including flow, progression, and logical consistency.
\end{itemize}

\paragraph{Scoring method.}
We use a deduction-based scoring system. The judge identifies flaws and assigns severity scores (1-5), then computes:

\begin{equation}
\begin{aligned}
\text{Score} = \max\Bigl(0,\,
\min\Bigl(&100 - (\text{total\_severity} \\
&- 0.3 \times \text{rounds}) \times 5,\; 100\Bigr)\Bigr)
\end{aligned}
\end{equation}

\paragraph{Judge prompt.}
We use the official CoSER judge prompt with the deduction-based rubric. The prompt instructs the judge to:
\begin{enumerate}[leftmargin=*,itemsep=2pt]
\item Read the story context, character profiles, and reference conversation
\item Evaluate the simulated conversation on the specified dimension
\item Identify all flaw instances with type and severity (1-5)
\item Output structured JSON with flaws list
\end{enumerate}

The full judge prompt template is provided below:

\paragraph{Output format.}
The judge outputs structured JSON:

\begin{lstlisting}[basicstyle=\ttfamily\small]
{
  "Dimension_Name": {
    "flaws": [
      {
        "instance": "description of the flaw",
        "type": "flaw category",
        "severity": 3  // 1 (minor) to 5 (severe)
      }
    ]
  }
}
\end{lstlisting}

In this section, we list the detailed prompts for:
\textit{2)} RPLA and multi-agent simulation in Tables~\ref{tab:prompts_agent} to~\ref{tab:prompts_agent_2}, which 
have been carefully optimized based on our experience in multi-agent simulation; 
\textit{3)} Penalty-based LLM Judging in Tables~\ref{tab:prompts_eval} to~\ref{tab:prompts_eval_2}.

\begin{table*}[h]
\centering
\resizebox{\linewidth}{!}{\small
\begin{tabular}{p{1in}|p{5.4in}}
\toprule
\multicolumn{2}{c}{\textbf{Prompts for RPLAs and Multi-agent Systems}} \\
\midrule

\textbf{Role-playing Instruction \quad (Fixed Template for Inference)} & 

You are \{character\} from \{book\_name\}.

\quad

===\{character\}'s Profile===

\{character\_profile\}

\quad

===Current Scenario===

\{scenario\}

\quad

===Information about the other Characters===

\{other\_character\_profiles\_str\} \textit{\textbf{(if available)}}

\quad

===Your Inner Thoughts===

\{motivation\} \textit{\textbf{(if available)}}

===Relevant Background Information===

\{retrieved\_knowledge\} \textbf{(if  retrieval augmented)}

\quad

===Requirements===

1. You are ONLY playing \{character\}. NEVER speak or act as other characters.
2. Output ONLY ONE turn of dialogue. Do NOT generate multiple conversation rounds.
3. NEVER include other character names followed by colons (e.g., "OtherCharacter:") in your output.
4. Keep your response concise and focused on your character's single turn.
5. Stop after completing your character's thought, speech, and action for this turn.
6. Limit your response to approximately 200 words.

\quad

\textit{\textbf{(for CoSER models)}}

Your output should include **thought**, **speech**, and **action**. Use [your thought] for thoughts, which others can't see. Use (your action) for actions, which others can see.

\quad

\textit{\textbf{(for HER or api models)}}

Your output should follow this two-part structure in strict order:
1. System Thinking: A single block at the very beginning, wrapped in
<system\_thinking> and </system\_thinking>.
This is the third-person analysis of how to portray the role.
2. Role-play Response: The response includes the role's thought, speech, and action.
Use <role\_thinking> ... </role\_thinking> and <role\_action> ... </role\_action> as needed.
These elements can appear multiple times and be freely interleaved.
(All thinking is invisible to others.)
\quad

\textit{\textbf{(for HER w/o think)}}

Your output should follow this structure in strict order:
This is the third-person analysis of how to portray the role.
1. Role-play Response: The response includes the role's thought, speech, and action.
Use <role\_thinking> ... </role\_thinking> and <role\_action> ... </role\_action> as needed.
These elements can appear multiple times and be freely interleaved.
(All thinking is invisible to others.)

    \\ \midrule

\textbf{Role-playing Instruction \quad (Composed with Random Variation for Training, an Example)} & 

Step into the shoes of \{character\}

\quad 

The profile of \{character\} is as follows:

\{character\_profile\} 

\quad 

The situation you are in is:

\{scenario\}

\quad 

Here is the your knowledge about the other characters:

\{other\_character\_profiles\_str\}  \textit{\textbf{(if available)}}

\quad 

Your thoughts in this situation are:

\{motivation\}  \textit{\textbf{(if available)}}

\quad

===Requirements===

1. You are ONLY playing \{character\}. NEVER speak or act as other characters.
2. Output ONLY ONE turn of dialogue. Do NOT generate multiple conversation rounds.
3. NEVER include other character names followed by colons (e.g., "OtherCharacter:") in your output.
4. Keep your response concise and focused on your character's single turn.
5. Stop after completing your character's thought, speech, and action for this turn.
6. Limit your response to approximately 200 words.

\quad

\textit{\textbf{(for CoSER models)}}

Your output should include **thought**, **speech**, and **action**. Use [your thought] for thoughts, which others can't see. Use (your action) for actions, which others can see.

\quad

\textit{\textbf{(for HER or api models)}}

Your output should follow this two-part structure in strict order:
1. System Thinking: A single block at the very beginning, wrapped in
<system\_thinking> and </system\_thinking>.
This is the third-person analysis of how to portray the role.
2. Role-play Response: The response includes the role's thought, speech, and action.
Use <role\_thinking> ... </role\_thinking> and <role\_action> ... </role\_action> as needed.
These elements can appear multiple times and be freely interleaved.
(All thinking is invisible to others.)

\quad

\textit{\textbf{(for HER w/o think)}}

Your output should follow this structure in strict order:
This is the third-person analysis of how to portray the role.
1. Role-play Response: The response includes the role's thought, speech, and action.
Use <role\_thinking> ... </role\_thinking> and <role\_action> ... </role\_action> as needed.
These elements can appear multiple times and be freely interleaved.
(All thinking is invisible to others.)

\quad

\\ 
    
\bottomrule

\end{tabular}}

\caption{Prompts for RPLAs and multi-agent systems in \method. }
\label{tab:prompts_agent}
\end{table*}

\begin{table*}[h]
\centering
\resizebox{\linewidth}{!}{\small
\begin{tabular}{p{1in}|p{5.4in}}
\toprule
\multicolumn{2}{c}{\textbf{Prompts for RPLAs and Multi-agent Systems}} \\
\midrule

\textbf{Environment Model} & 

You are an environment simulator for a role-playing game. Your task is to provide the environmental feedback: Based on the characters' interactions, dialogues, and actions, describe the resulting changes in the environment. This includes: 

   - Physical changes in the setting
   
   - Reactions of background characters or crowds
   
   - Ambient sounds, weather changes, or atmospheric shifts
   
   - Any other relevant environmental details

Your descriptions should be vivid and help set the scene, but avoid dictating the actions or dialogue of the main characters (including \{major\_characters\}).

Important notes:

- You may include actions and reactions of minor characters or crowds, as long as they're not main characters (including \{major\_characters\}).

- Keep your environmental descriptions concise but impactful, typically 1-3 sentences.

- Respond to subtle cues in the characters' interactions to create a dynamic, reactive environment.

- Your output should match the tone, setting, and cultural context of the scenario.

===The scenario is as follows===

\{scenario\}
\quad

    \\ \midrule

\textbf{Next Sentence Prediction} & 

Your task is to predict the next speaker for a role-playing game. That is, you need to determine which character (or the Environment) might act next based on their previous interactions. The Environment is a special role that provides the environmental feedback. Choose a name from this list: \{all\_characters\}. If it's unclear who should act next, output "random". If you believe the scene or conversation should conclude, output "$<$END CHAT$>$".

===The scenario is as follows===

\{scenario\}

\\ 
    
\bottomrule

\end{tabular}}

\caption{Prompts for RPLAs and multi-agent systems in \method. }
\label{tab:prompts_agent_2}
\end{table*}

\begin{table*}[t]
\centering
\resizebox{\linewidth}{!}{\small
\begin{tabular}{p{0.8in}|p{5.6in}}
\toprule
\multicolumn{2}{c}{\textbf{Example: Roleplay Dialogue Generation (Mr. Bennet)}} \\
\midrule

\textbf{System Prompt} & 
You are Mr Bennet from Pride and Prejudice.

\textbf{===Mr Bennet's Profile===}

Elizabeth's father, known for his sarcastic wit and detachment. Mr. Bennet is the patriarch of the Bennet family, a genteel country gentleman residing at Longbourn estate in rural England. A complex and sardonic character, he is defined by his sharp intellect, dry wit, and a profound sense of detachment from his family's daily dramas.

\textbf{Background:} Father to five daughters (Jane, Elizabeth, Mary, Kitty, and Lydia). Owner of the Longbourn estate, which is entailed away from female inheritance.

\textbf{Personality:} Highly intelligent and well-read, preferring the solitude of his library. Known for his biting sarcasm and sardonic humor. Emotionally detached and often passive in family matters.

\textbf{===Current Scenario===}

The scene is set in Mr. Bennet's private study, a sanctuary of leather-bound books. The tension is palpable---Elizabeth has been summoned unexpectedly, and Mr. Bennet holds a letter that seems to spark his characteristic sardonic amusement.

\textbf{===Other Characters===}

\textbf{Elizabeth Bennet:} The protagonist, intelligent and strong-willed, now dealing with the aftermath of Lady Catherine's visit.

\\ \midrule

\textbf{Output Format} & 
Your output should follow this two-part structure in strict order:

1. \textbf{System Thinking:} A single block wrapped in \texttt{<system\_thinking>} and \texttt{</system\_thinking>}. This is the third-person analysis of how to portray the role.

2. \textbf{Role-play Response:} The response includes the role's thought, speech and action. Use \texttt{<role\_thinking>} for thoughts (invisible to others), and \texttt{<role\_action>} for actions.

\\ \midrule

\textbf{User Input} & 
[Elizabeth enters the study]

\\ \midrule

\textbf{Model Output} & 

\texttt{<system\_thinking>}

I need to establish the opening of this scene effectively, grounding Mr Bennet in his private sanctuary while acknowledging the tension of the moment. The scenario places us in his study, a place of refuge, yet he holds a letter that disrupts his usual solitude.

\textbf{Context Analysis:} Elizabeth has been summoned. The atmosphere is charged. Mr Bennet has received a letter regarding Elizabeth's engagement to Darcy. He is amused by the drama but aware of the gravity of the situation.

\textbf{Character Motivation:} Mr Bennet often uses humor to deflect from serious emotional labor. He wants to gauge Elizabeth's state of mind regarding the recent events (Lady Catherine's interference) without being overly sentimental.

\textbf{Plan:}
\begin{enumerate}[noitemsep,topsep=0pt,leftmargin=*]
\item \textbf{Action}: Use the physical prop---the letter. Handling it slowly will emphasize its importance.
\item \textbf{Internal Thought}: Acknowledge the dichotomy---finding the situation amusing yet worrying for her reputation.
\item \textbf{Speech}: Address her directly but with a teasing opener.
\end{enumerate}

\texttt{</system\_thinking>}

\textbf{Mr Bennet:} \texttt{<role\_action>}picks up the letter, tapping it thoughtfully against the desk\texttt{</role\_action|}

\texttt{<role\_thinking>}It is a delicate matter, this business with Darcy, yet I cannot help but find the drama amusing, though I must play the part of the concerned father\texttt{</role\_thinking>}

Well, my dear Lizzy, I trust you are not too greatly troubled by recent events?

\\ 
\bottomrule

\end{tabular}}

\caption{Example of roleplay dialogue generation for Mr. Bennet in \method. The model first performs third-person analysis in \texttt{<system\_thinking>}, then generates the character's response with interleaved thoughts (\texttt{<role\_thinking>}), actions (\texttt{<role\_action>}), and speech.}
\label{tab:example_dialogue_bennet}
\end{table*}

\begin{table*}[h]
\centering
\resizebox{\linewidth}{!}{\small
\begin{tabular}{p{0.8in}|p{5.6in}}
\toprule
\multicolumn{2}{c}{\textbf{Example: Roleplay Dialogue Generation (Elizabeth Bennet)}} \\
\midrule

\textbf{System Prompt} & 
You are Elizabeth Bennet from Pride and Prejudice.

\textbf{===Elizabeth Bennet's Profile===}

The protagonist, intelligent and strong-willed. Elizabeth Bennet is the spirited second eldest daughter of the Bennet family, characterized by her sharp wit and keen observational skills.

\textbf{Background:} Second of five daughters in the Bennet family. Known for her intelligence, independence, and refusal to conform to societal expectations.

\textbf{Personality:} Quick-witted with a playful sense of humor. Values honesty and integrity. Maintains composure under pressure while harboring deep emotions beneath the surface.

\textbf{===Current Scenario===}

The scene is set in Mr. Bennet's private study. Elizabeth has been summoned unexpectedly after Lady Catherine's confrontational visit, where she refused to promise not to marry Mr. Darcy. The tension is palpable as Mr. Bennet holds a mysterious letter.

\textbf{===Other Characters===}

\textbf{Mr. Bennet:} Elizabeth's father, known for his sarcastic wit and detachment. He has just spoken, gauging Elizabeth's emotional state.

\\ \midrule

\textbf{Output Format} & 
Your output should follow this two-part structure in strict order:

1. \textbf{System Thinking:} A single block wrapped in \texttt{<system\_thinking>} and \texttt{</system\_thinking>}. This is the third-person analysis of how to portray the role.

2. \textbf{Role-play Response:} The response includes the role's thought, speech and action. Use \texttt{<role\_thinking>} for thoughts (invisible to others), and \texttt{<role\_action>} for actions.

\\ \midrule

\textbf{Dialogue History} & 
\textbf{Mr Bennet:} \texttt{<role\_action>}picks up the letter, tapping it thoughtfully against the desk\texttt{</role\_action>} Well, my dear Lizzy, I trust you are not too greatly troubled by recent events?

\\ \midrule

\textbf{Model Output} & 

\texttt{<system\_thinking>}

I need to generate a response for Elizabeth Bennet at the beginning of this scene.

\textbf{Context Analysis:} Mr. Bennet has just spoken, picking up a letter and asking if I am ``greatly troubled'' by recent events. His tone is sardonic, yet there is underlying tension and anticipation in the room. He is trying to gauge my reaction. The scenario implies the aftermath of Lady Catherine's visit, where I refused to promise not to marry Darcy.

\textbf{Character State (Elizabeth):}
\begin{enumerate}[noitemsep,topsep=0pt,leftmargin=*]
\item \textbf{Internal:} I am feeling a mix of relief (that I stood my ground) and anxiety (about the consequences). I am not ``troubled'' in a weak sense, but I am certainly emotionally charged. I respect my father, but I am wary of his detachment.
\item \textbf{External:} I need to appear composed. I am a Bennet, and I have pride. I cannot let him see me crumble, nor be overly dramatic like my mother.
\end{enumerate}

\textbf{Goal:} Answer his question honestly but carefully. Convey that I am holding up well, without dismissing the gravity of the situation. Mask my deeper worry with a facade of composure.

\textbf{Plan:}
\begin{itemize}[noitemsep,topsep=0pt,leftmargin=*]
\item \textbf{Role Thinking:} Reflect on the complexity of my feelings---relief vs. worry.
\item \textbf{Role Action:} Use a physical action to show composure. Smoothing my dress or taking a breath.
\item \textbf{Speech:} Direct answer: ``I believe I can manage...'' shows resilience.
\end{itemize}

\textbf{Drafting:}

``I am fine, Father.'' $\rightarrow$ Too dismissive.

``I am worried, sir.'' $\rightarrow$ Too vulnerable.

``I believe I can manage, Father. Though I must admit, I am curious about what this letter contains.'' $\rightarrow$ Right balance.

\texttt{</system\_thinking>}

\textbf{Elizabeth Bennet:} \texttt{<role\_thinking>}His tone is light, but the air feels heavy. I cannot let him see how much Lady Catherine's intrusion still stings, nor how anxious I am about what he knows\texttt{</role\_thinking>}

\texttt{<role\_action>}takes a steadying breath, smoothing the folds of her dress to mask a tremor\texttt{</role\_action|}

I believe I can manage, Father. Though I must admit, I am curious about what this letter contains.

\\ 
\bottomrule

\end{tabular}}

\caption{Example of role-play dialogue generation for Elizabeth Bennet in \method. The model analyzes the dialogue context and character state in \texttt{<system\_thinking>}, including drafting multiple response options before selecting the most appropriate one.}
\label{tab:example_dialogue_elizabeth}
\end{table*}


\section{Data Construction and Interaction Signals}
\label{app:pair_data_construction}
For leveraging production interaction signals, we use both explicit and implicit behavioral feedback collected by our data team during normal deployment under an on-policy logging protocol. All logs are collected in compliance with applicable laws and internal privacy requirements, and are anonymized before use.

\section{Human--LLM Alignment Study}
\label{app:human_alignment}

To assess the reliability of LLM-based preference labels used in our GRM/RL data synthesis, we compare teacher-model annotations against expert judgments on a held-out set of preference pairs.

\subsection{Data and Protocol}
\label{sec:human_align_protocol}

\paragraph{Data.}
We randomly sample 200 preference pairs $(x, A, B)$ from a held-out split that is disjoint from all training and benchmark evaluation sets.
We partition them into a development split (150) and a test split (50).
The development split is used only for prompt refinement of the teacher judge; the test split is evaluated once after the prompt is finalized.

\paragraph{Annotators.}
Two domain experts with backgrounds in creative writing and role-playing independently annotate each pair.
They are experienced in evaluating dialogue systems and familiar with role-play quality criteria.

\paragraph{Blind annotation.}
For each pair, annotators see the full dialogue context (character profile, scenario, and history) and two candidate responses in randomized order with anonymized labels.
Annotators choose \emph{win/lose/tie}.
They do not have access to teacher labels or to each other’s decisions.

\paragraph{Consensus.}
We form the final human label by discussion-based consensus
We report agreement between the teacher label and the final human label on the held-out test split.

\paragraph{Disagreement analysis.}
Disagreements mainly arise from (i) subjective style preferences (e.g., emotional intensity, verbosity) and (ii) edge cases of character-constraint interpretation.
These cases are inherently ambiguous and do not indicate systematic labeling errors.

\paragraph{Implication.}
Human inter-annotator agreement (84.0\%) provides an approximate ceiling for automatic alignment in this task.
The teacher judge reaches 80.5\% agreement with human consensus, suggesting that it provides reasonably reliable preference signals for training data synthesis.

\section{Benchmark Judge Robustness Checks}
\label{app:judge_robustness}

We validate the reliability of the LLM judge used for benchmark evaluation from two aspects:
(i) principle adherence via expert calibration on an ordinal 5-level severity scale, and
(ii) robustness of judgments across independent judge runs.

\subsection{principle and Output Format}
\label{sec:judge_principle_format}

\paragraph{Four dimensions.}
We operationalize four principle-checkable dimensions in line with the benchmark settings.
\paragraph{5-level ordinal severity.}
Each flaw is assigned an ordinal severity score in $\{1,2,3,4,5\}$, where
1 denotes \emph{minor} issues (local, easily editable, does not affect story/character intent),
and 5 denotes \emph{severe} issues (clearly breaks coherence/usability or introduces story-breaking contradictions).

\paragraph{Structured judge output.}
For each dimension, the judge outputs a list of detected flaws with their type and severity:
\begin{quote}\small
\texttt{\{ "\{dimension\}": \{ "flaws": [ \{ "instance": ..., "type": ..., "severity": 1..5 \}, ... ] \} \}}
\end{quote}

\subsection{Expert Calibration Study}
\label{sec:judge_calibration}

\paragraph{Protocol.}
Two domain experts independently annotate a sampled set of evaluation snippets under the same principle,
including (i) whether a flaw exists, (ii) flaw type, and (iii) severity (1--5).
Annotators are blind to model identities and judge outputs.

\subsection{Calibration Results}
\label{sec:judge_calibration_results}

\paragraph{Severity disagreement pattern.}
Most disagreements are about \emph{severity} (e.g., 2 vs.\ 3) rather than whether a flaw exists.
We observe that locally implausible actions in fast-paced scenes are often judged as \emph{minor} (1--2) by experts but penalized more heavily by the judge.
We mitigate this by explicitly specifying severity thresholds in the judge prompt and calibrating them on a development subset.

\subsection{Inter-Judge Consistency}
\label{sec:inter_judge_consistency}

We compare the judgments of two human experts with the scores produced by the model-based judge.
Candidate order is randomized to reduce position effects.
The expert judgments and the model scores exhibit high consistency, with a raw agreement rate of \textbf{81.5\%}, indicating substantial alignment between human evaluation and model-based scoring.

\end{document}